\documentclass[conference,a4paper]{IEEEtran}
\IEEEoverridecommandlockouts
\usepackage{amsmath,amssymb}
\usepackage{cite}

\usepackage{booktabs}
\usepackage{graphicx}
\usepackage{url}
\usepackage[table]{xcolor}
\def\BibTeX{{\rm B\kern-.05em{\sc i\kern-.025em b}\kern-.08em
    T\kern-.1667em\lower.7ex\hbox{E}\kern-.125emX}}
\usepackage{subcaption}
\usepackage{algorithm}
\usepackage{algpseudocode}
\usepackage{multirow}
\def\dif{\mathop{}\hphantom{\mskip-\thinmuskip}\mathrm{d}}%
\let\daccent\d
\let\d\relax
\newcommand\d{\ifmmode\dif\else\expandafter\daccent\fi}

\definecolor{CoralRed}{RGB}{255, 107, 77}   % 橙红色
\definecolor{Apricot}{RGB}{255, 228, 181} % 杏色

\begin{document}

\title{Low-Frequency First: Eliminating Floating Artifacts in 3D Gaussian Splatting}

% \author{Anonymous Authors}
\author{
\IEEEauthorblockN{Jianchao Wang, Peng Zhou, Cen Li, Rong Quan, Jie Qin\IEEEauthorrefmark{1}\thanks{\IEEEauthorrefmark{1}Corresponding author.}}
\IEEEauthorblockA{
    College of Artificial Intelligence, Nanjing University of Aeronautics and Astronautics \\
    Key Laboratory of Brain-Machine Intelligence Technology, Ministry of Education, China\\
    wangjianchao@nuaa.edu.cn
}
}

\maketitle
\renewcommand{\thefootnote}{}
\footnote{
This work was partially supported by the National Natural Science Foundation of China (No. 62276129 \& 62206127), the Fundamental Research Funds for the Central Universities (No. NS2024060), the Natural Science Foundation of Jiangsu Province (No. BK20241400) and the China National Postdoctoral Program for Innovative Talents (No. BX20240484).
}
\begin{abstract}
3D Gaussian Splatting (3DGS) is a powerful and computationally efficient representation for 3D reconstruction.  Despite its strengths, 3DGS often produces floating artifacts, which are erroneous structures detached from the actual geometry and significantly degrade visual fidelity. The underlying mechanisms causing these artifacts, particularly in low-quality initialization scenarios, have not been fully explored. In this paper, we investigate the origins of floating artifacts from a frequency-domain perspective and identify under-optimized Gaussians as the primary source. Based on our analysis, we propose \textit{Eliminating-Floating-Artifacts} Gaussian Splatting (EFA-GS), which selectively expands under-optimized Gaussians to prioritize accurate low-frequency learning. Additionally, we introduce complementary depth-based and scale-based strategies to dynamically refine Gaussian expansion, effectively mitigating detail erosion. Extensive experiments on both synthetic and real-world datasets demonstrate that EFA-GS substantially reduces floating artifacts while preserving high-frequency details, achieving an improvement of 1.68 dB in PSNR over baseline method on our RWLQ dataset. Furthermore, we validate the effectiveness of our approach in downstream 3D editing tasks. We provide our implementation in https://jcwang-gh.github.io/EFA-GS.
\end{abstract}


\section{Introduction}
\label{sec:intro}

3D Gaussian Splatting (3DGS)~\cite{kerbl3Dgaussians} is an explicit 3D representation approach that combines differentiable rendering with explicit representations, achieving efficient and high-quality novel view synthesis.
Due to its highly qualified real-time performance, 3DGS has been extended to various applications: volumetric video reconstruction~\cite{xu2024representing}, 3D generation~\cite{tang2023dreamgaussian}, 3D editing~\cite{chen2024gaussianeditor,wang2024gaussianeditor}, etc. Although 3DGS achieves outstanding results on various tasks, it is highly dependent on the quality of 3D object/scene datasets~\cite{trackgs,Jiang2025icra,Fu_2024_CVPR}. 
% Specifically, 3DGS may underperform on low-quality datasets (with inaccurate initialization or low-quality images), as illustrated in Fig.~\ref{fig:demofig}. One factor leading to poor performance is the existence of floating artifacts.
However, on low-quality datasets with inaccurate initialization or low-resolution images, 3DGS often suffers from visual artifacts that degrade its overall performance.
% as shown in Fig.~\ref{fig:demofig}.

\begin{figure}[htbp]
  \centering
  \includegraphics[width=\linewidth]{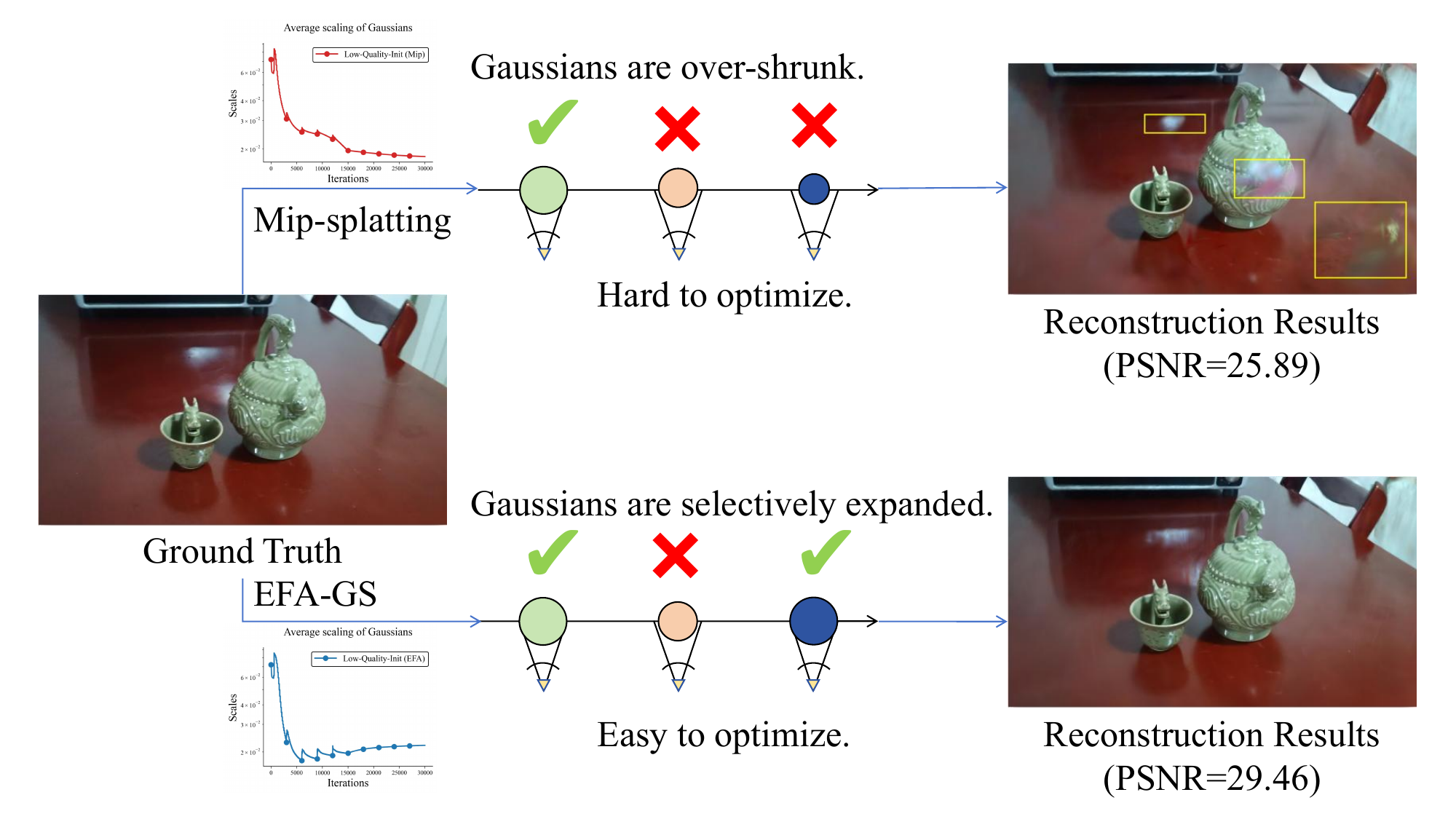}
  \caption{Illustration of our EFA-GS. Ordinary 3DGS frameworks (such as Mip-splatting~\cite{yu2024mip}) sometimes have a frequency bias in the training process and low-quality initialization exacerbate this phenomenon, resulting in more over-shrunk Gaussians. Our EFA-GS successfully mitigate this issue and improve the performance by selectively expanding Gaussians.}
  \label{fig:demofig}
\end{figure}

\begin{figure}[htbp]
  \centering
  \begin{subfigure}[t]{0.23\linewidth}
    \includegraphics[width=\linewidth]{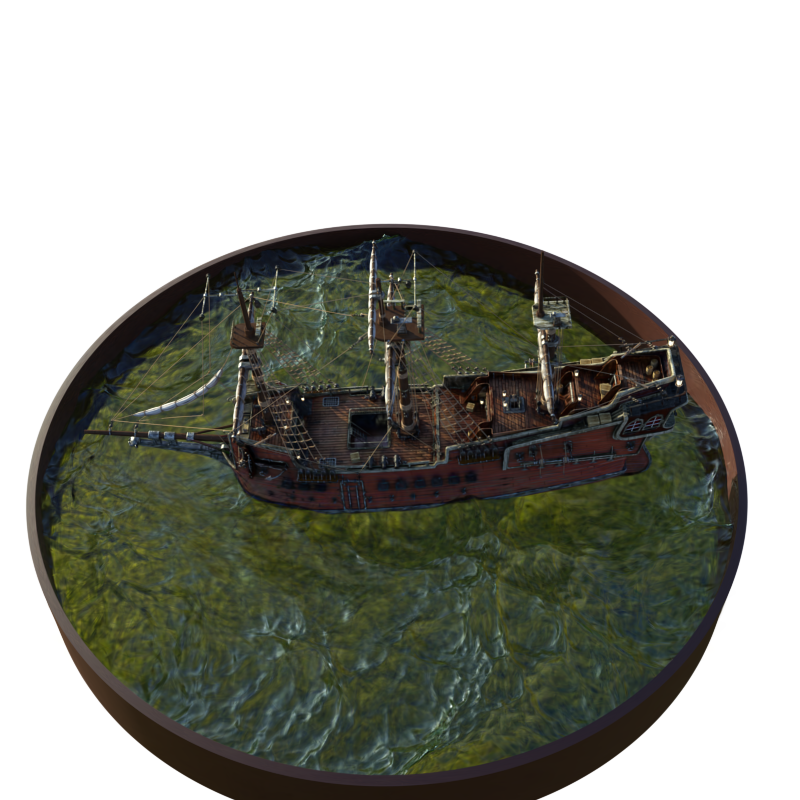}
    \includegraphics[width=\linewidth]{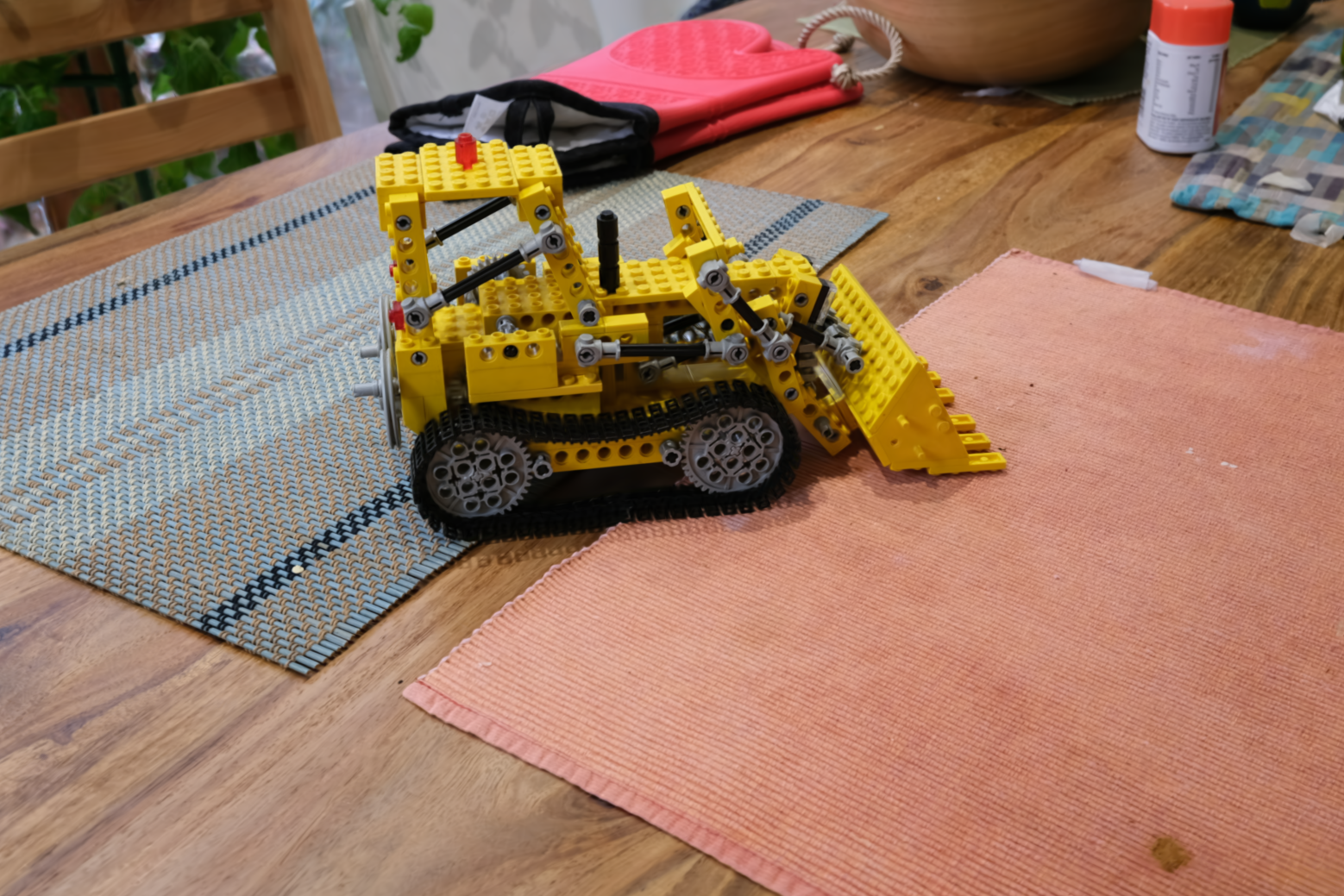}
    \caption{Clean init.}
  \end{subfigure}
  \begin{subfigure}[t]{0.23\linewidth}
    \includegraphics[width=\linewidth]{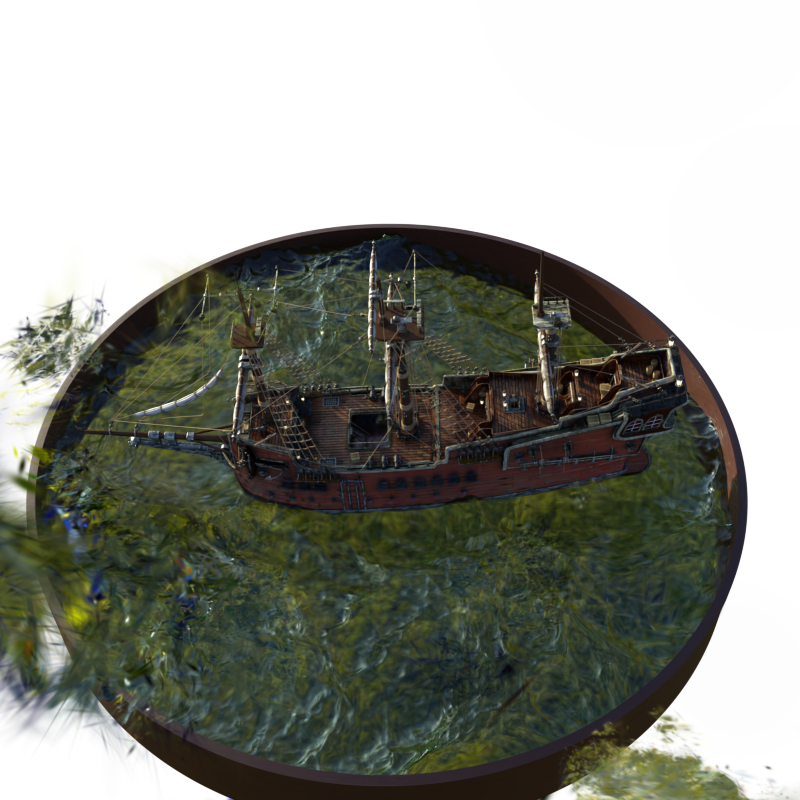}
    \includegraphics[width=\linewidth]{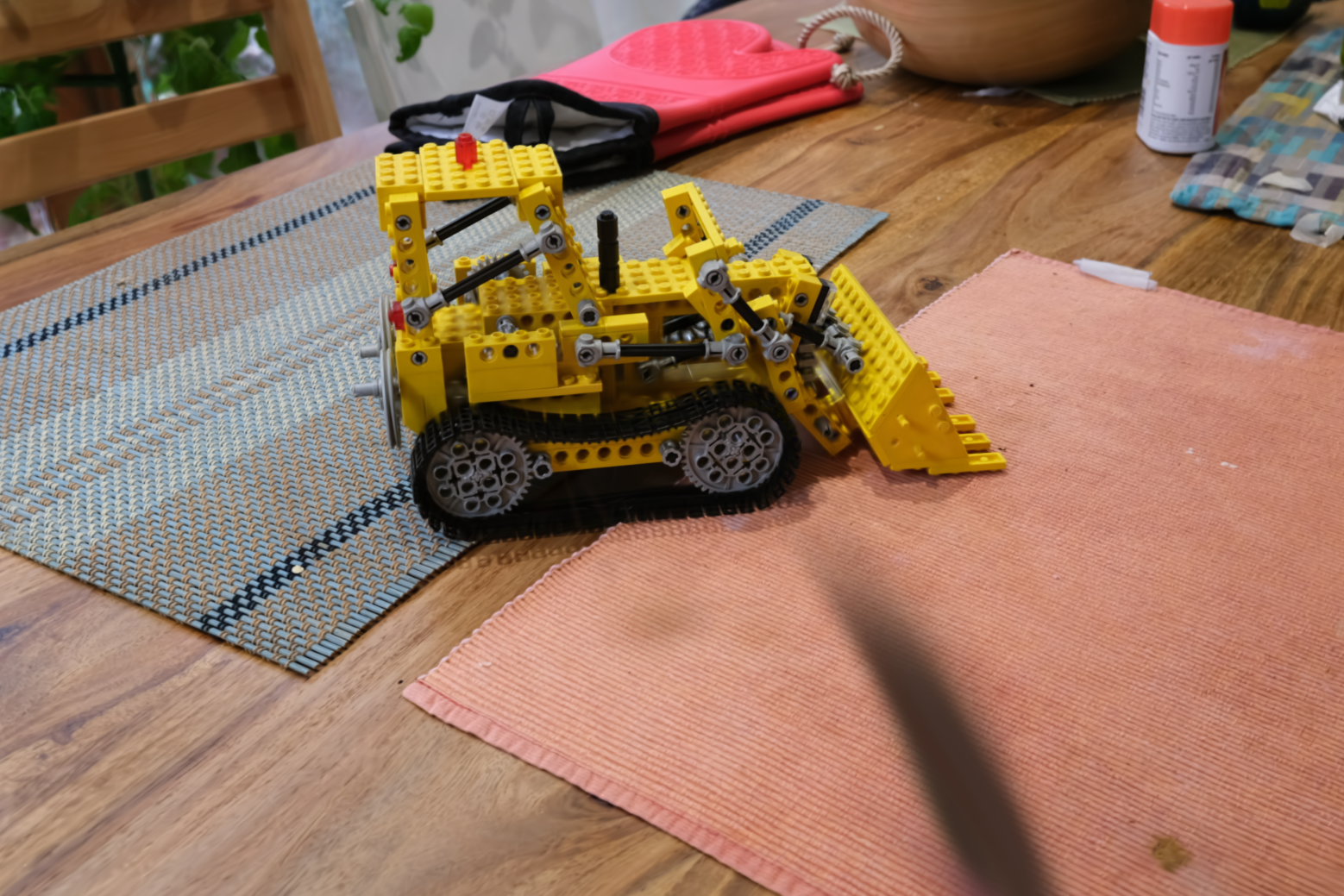}
    \caption{Noisy init.}
  \end{subfigure}
  \begin{subfigure}[t]{0.23\linewidth}
    \includegraphics[width=\linewidth]{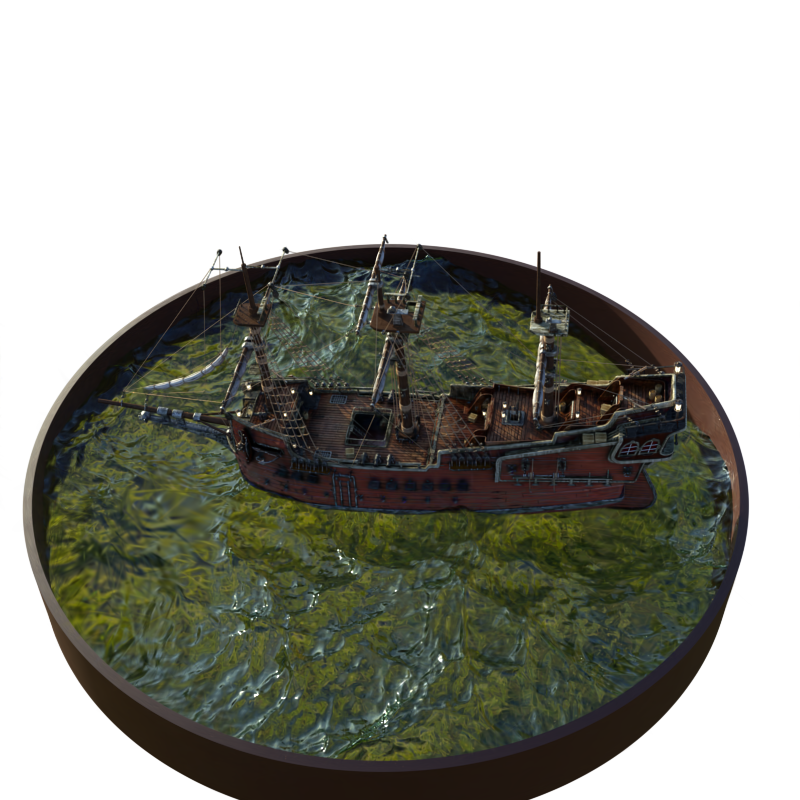}
    \includegraphics[width=\linewidth]{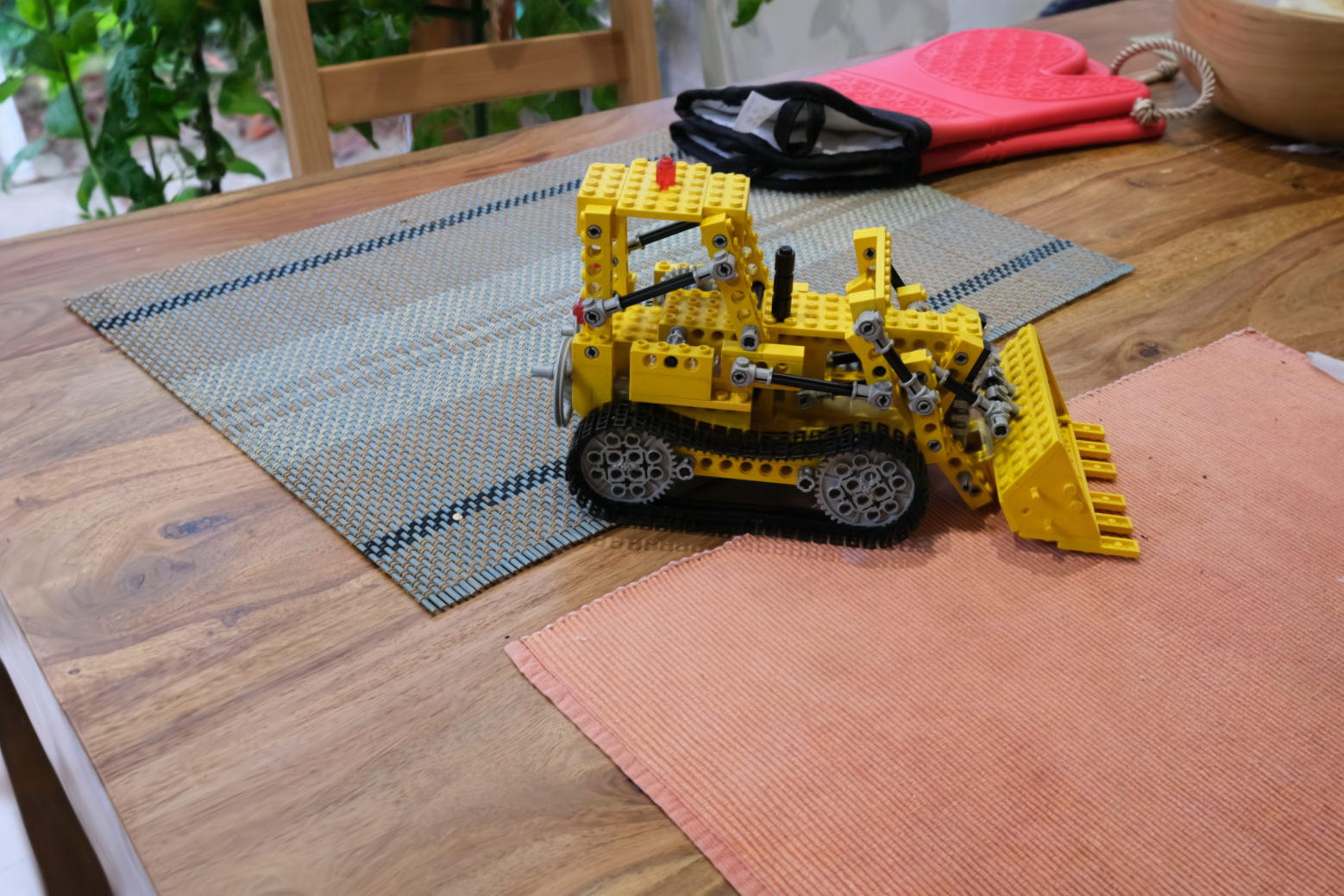}
    \caption{Noisy init (training).}
  \end{subfigure}
  \begin{subfigure}[t]{0.23\linewidth}
    \includegraphics[width=\linewidth]{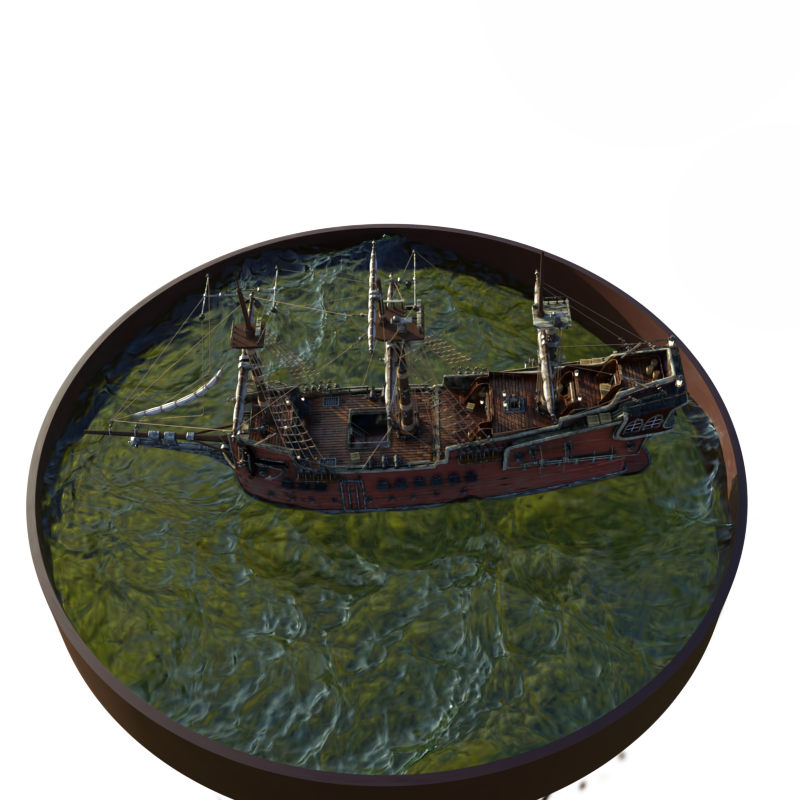}
    \includegraphics[width=\linewidth]{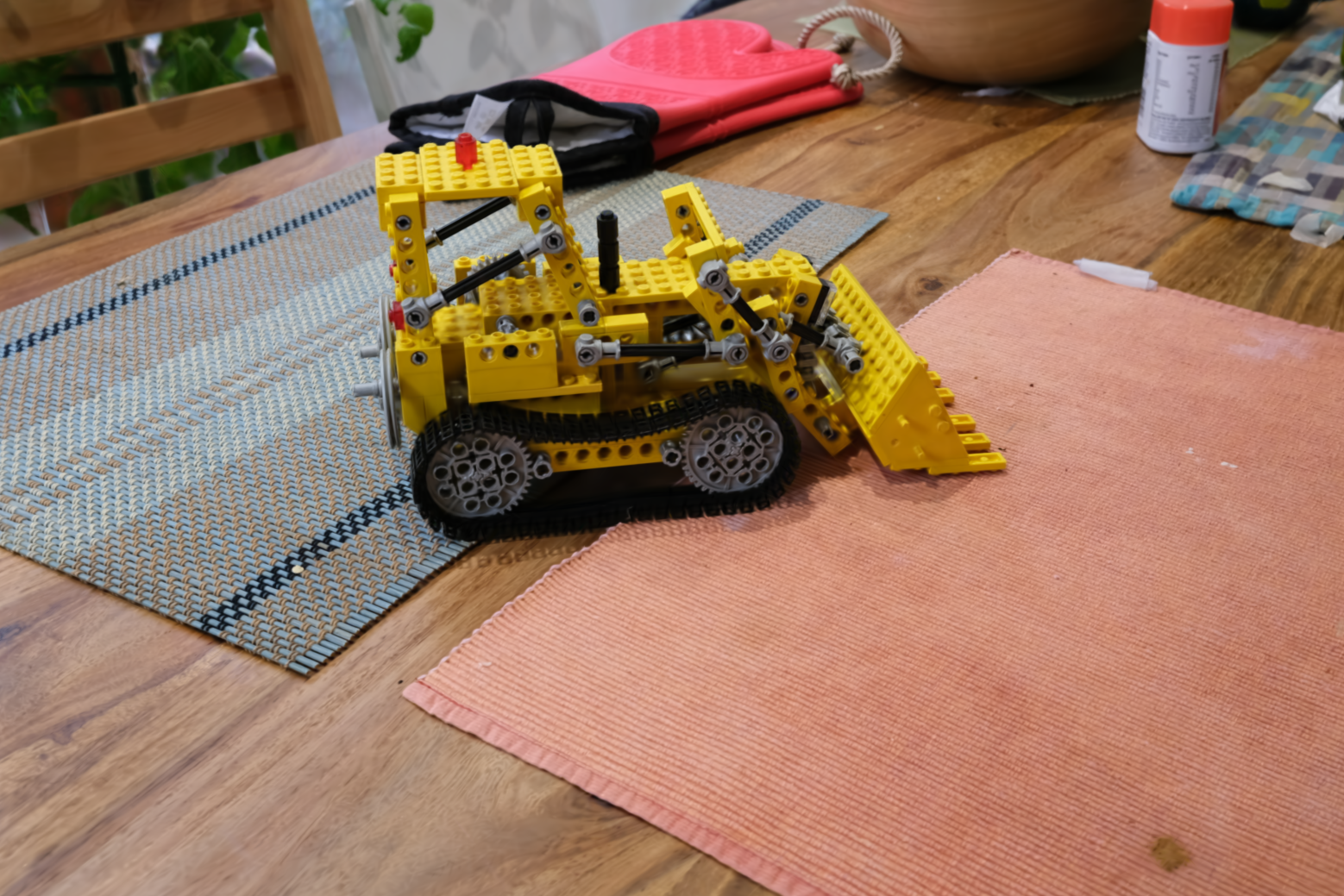}
    \caption{Noisy init with EFA-GS.}
  \end{subfigure}
  \caption{We train Mip-splatting models~\cite{yu2024mip} on ship and kitchen~\cite{mildenhall2020nerf,barron2022mip} for 30k iterations and results are shown in (a), (b) and (c). We inject noise into Gaussian scales of ship and Gaussian coordinates of kitchen respectively. (a),(b) and (d) are the rendered images from testing view whereas (c) are the rendered images from training view(close to the testing view). Obviously, noisy initialization introduces floating artifacts, and (d) demonstrates that EFA-GS (simple) effectively eliminates these floating artifacts.}
  \label{fig:noisyexp}
\end{figure}

% Floating artifacts are sometimes observed in 3DGS-based reconstruction and editing tasks. 
% These artifacts manifest as ghostly structures spatially detached from the true 3D geometry, significantly degrade rendering fidelity. 
As illustrated in Fig.~\ref{fig:demofig}, a typical failure in low-quality scenarios is the appearance of floating artifacts, visible as structures detached from the actual geometry.
Our analysis reveals that low-quality initialization is a primary cause to floating artifacts, as illustrated in Fig.~\ref{fig:noisyexp} (a) and (b). 
To investigate this phenomenon further, we employ frequency analysis (Sec.~\ref{sec:theanal}), building upon the sampling rate framework introduced by Yu~\cite{yu2024mip}. Specifically, we examine the relationship between frequency components and Gaussian covariance matrices.
Through experiments, we identify the key factor contributing to floating artifacts and provide a theoretical justification to explain our findings: Gaussians smaller than the Nyquist sampling interval are probably under-optimized and exhibit reduced low-frequency information. 
Furthermore, noise in the initialization further amplifies the problem by introducing more under-optimized Gaussians that fail to capture reliable low-frequency content, leading to visible floating artifacts.
% noise in initialization priors exacerbates this issue, generating additional under-optimized Gaussians that ultimately develop into floating artifacts due to the lack of accurate low-frequency information.

A low-pass filter is necessary to deal with floating artifacts. 
Despite employing low-pass filters, Mip-splatting~\cite{yu2024mip}, a widely adopted anti-aliasing technique in 3DGS, fails to eliminate most floating artifacts (Fig.~\ref{fig:noisyexp} (b)). 
We find that a frequency bias exists during training 3DGS: The Gaussian scaling dynamics exhibits a strikingly rapid collapse during early training—plummeting within merely ~5,000 iterations—before transitioning to a slow decay phase (Fig.~\ref{fig:numofgauss} (c)). 
This highly non-stationary scaling behavior, further exacerbated by noise, cannot be adequately addressed by static low-pass filtering. Moreover, over-optimized Gaussians and under-optimized Gaussians should be treated differently, whereas Mip-splatting simply applies similar low-pass filters to all Gaussians.

To overcome this limitation, we propose to directly manipulate Gaussian frequencies. 
Specifically, we design the \textit{Low-Frequency-Come-First} (LFCF) algorithm. LFCF selects under-optimized Gaussians according to their cumulative gradient information and expands them to help them learn accurate low-frequency information.
Experiments indicate that LFCF effectively reduces most floating artifacts effectively.
While simply enforcing Gaussians to prioritize low frequencies helps eliminate floating artifacts, improper hyperparameter settings in LFCF can lead to undesirable blurring effects.
The trade-off may limit LFCF's generalizability across diverse scenarios.
To address this, we design a depth-based strategy and a scale-based strategy that dynamically assign optimal expansion parameters to each Gaussian. We also implement other coarse-to-fine techniques to preserve intricate details throughout the process. 

All of these designs are integrated into an improved 3DGS framework called \textit{Eliminating-Floating-Artifacts} Gaussian Splatting (EFA-GS). EFA-GS exhibits superior performance compared to the baseline method in experiments: As illustrated in Fig.~\ref{fig:demofig}, EFA-GS exceeds Mip-splatting by 3.57 dB in PSNR and effectively eliminate floating artifacts while maintaining details of the reflective area in the scene. Comprehensive experiments across both 3D reconstruction and editing tasks demonstrate that EFA-GS successfully addresses the critical challenge of floating artifact removal while maintaining superior detail preservation. 

In summary, the main contributions of this work are as follows:
\begin{itemize}
\item We perform a frequency-domain analysis that reveals the underlying mechanism of floating artifacts during the training of 3DGS. 
\item Based on this insight, we introduce EFA-GS, an enhanced 3DGS framework that mitigates floating artifacts while preserving fine-grained details in 3D reconstruction tasks.
\item We further demonstrate that EFA-GS is compatible with existing 3DGS-based editing pipelines, improving visual quality across a range of applications with limited computation costs.
\end{itemize}

\section{Related Works}
\label{sec:relat}

\subsection{3D Representations}

Traditional 3D representation methods have been widely used in all kinds of 3D tasks. 
As a MLP-based 3D representation model, NeRF~\cite{mildenhall2020nerf} is proposed in 2020 and leads a surge of 3D reconstruction/generation/editing tasks~\cite{barron2022mip,poole2022dreamfusion,Liu_2023_ICCV,Haque_2023_ICCV,Kuang_2023_CVPR}.
NeRF employs MLP and volume rendering to achieve delicate rendering results. 
However, training a NeRF is a time-consuming process due to MLP usage, which significantly prevents NeRF from being applied to various real-time 3D tasks~\cite{fridovich2022plenoxels,mueller2022instant}.

In recent years 3DGS was proposed to achieve high visual quality for real-time rendering tasks~\cite{kerbl3Dgaussians}. By integrating differential rendering and traditional explicit representation methods (point cloud), 3DGS has achieved state-of-the-art performance among real-time rendering methods and has been widely used in many other 3D tasks~\cite{Wu_2024_CVPR,Qin_2024_CVPR,Huang2DGS2024,guedon2024sugar,Matsuki2024gslam,yu2024mip}.

\begin{table}
    \caption{Reconstruction results of ship and kitchen using different initializations. Noisy initialization causes more damages on the rendering quality for testing views than training views, and expands the gap between training and testing views. Our EFA-GS (simple) effectively mitigate this issue.}
    \centering
    % \resizebox{0.48\textwidth}{!}{
        \begin{tabular}{c|cc|cc}
        \toprule
        \multirow{2}{*}{PSNR} & \multicolumn{2}{c}{ship~\cite{mildenhall2020nerf}} & \multicolumn{2}{c}{kitchen~\cite{barron2022mip}} \\
                                  & Train & Test  & Train & Test  \\
        \midrule
        Clean init (Mip-splatting)           & 33.94 & 29.74 & 34.74 & 32.03 \\
        Noisy init (Mip-splatting)           & 34.02 & 27.53 & 33.68 & 28.39 \\
        /Clean - Noisy/ (Mip-splatting)      &  0.08 &  2.21 &  1.06 &  3.64 \\
        \midrule
        Clean init (EFA-GS, simple)          & 32.88 & 30.66 & 31.54 & 31.23 \\
        Noisy init (EFA-GS, simple)          & 32.89 & 29.73 & 31.33 & 29.48 \\
        /Clean - Noisy/ (EFA-GS, simple)     &  0.01 &  0.93 &  0.21 &  1.75 \\
        \bottomrule
        \end{tabular}
    % }

    \label{tab:noisyexp}
\end{table}

\begin{figure*}[htbp]
  \centering
  \begin{subfigure}{0.3\linewidth}
    \includegraphics[width=\linewidth]{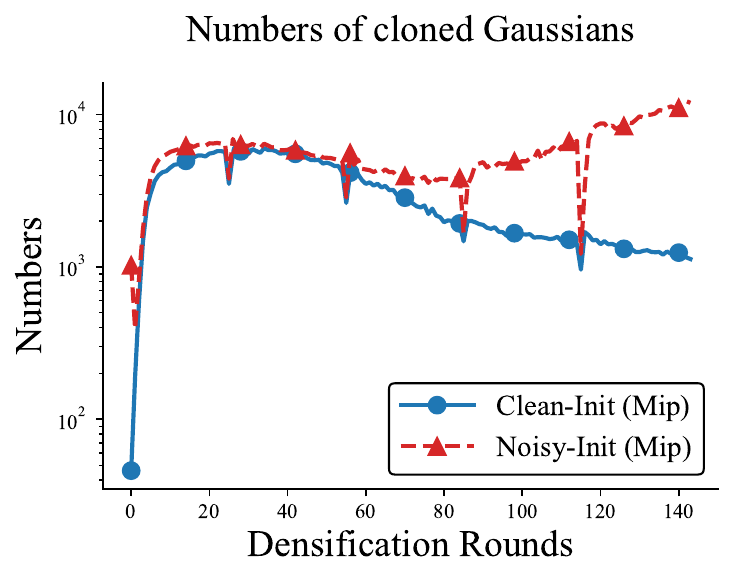}
    \caption{Numbers of cloned Gaussians.}
  \end{subfigure}
  \begin{subfigure}{0.3\linewidth}
    \includegraphics[width=\linewidth]{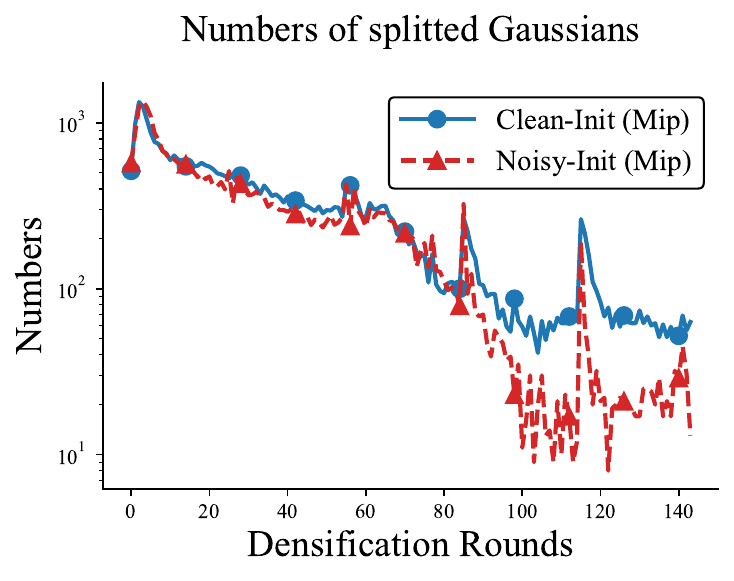}
    \caption{Numbers of splitted Gaussians.}
  \end{subfigure}
  \begin{subfigure}{0.3\linewidth}
    \includegraphics[width=\linewidth]{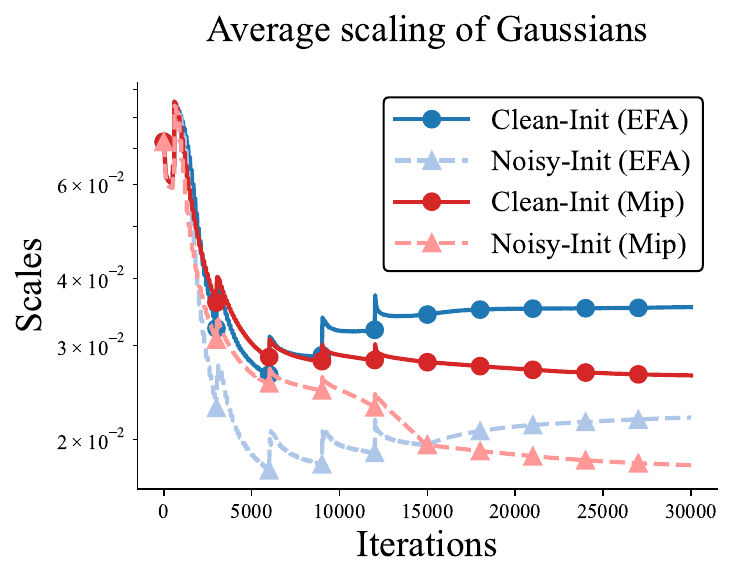}
    \caption{Average scalings of Gaussians.}
  \end{subfigure}
  \caption{We record the numbers of manipulated Gaussians of ship~\cite{mildenhall2020nerf} in the densification process(up to 15,000 iterations) and the average scaling of Gaussians in the whole optimization process. The results demonstrate that 3DGS also learns low frequency information first before diving into high frequency information. Furthermore, when noisy initialization were used, the model would split less Gaussians, clone more Gaussians, and most Gaussians become over-shrunk.}
  \label{fig:numofgauss}
\end{figure*}

However, 3DGS has some problems in practice, preventing it from being further applied to more 3D tasks. One of the issues is the existence of floating artifacts. Floating artifacts usually arise due to the insufficient quality of SfM initialization of the 3DGS model. To solve these issues, some methods have been proposed~\cite{Huang2DGS2024,sun2024hd,yu2024gaussian,liu20243dgs}. However, some of them employ neural networks or complicated losses, which can make training a time-consuming process; Huang \textit{et al.}~\cite{Huang2DGS2024,yu2024gaussian} propose depth-distortion loss to mitigate this issue, whereas our experiments reveal that depth-distortion loss sometimes erodes fine-grained details and does not eliminate artifacts effectively (Fig. 8 and 9 in Appendix. F).

Zhang~\cite{zhang2024fregs} employ low-pass and high-pass filters to extract different frequency components and design a frequency-based loss to address the over-reconstruction issue. Jung~\cite{jung2024relaxing} proposes Sparse-Large-Variance (SLV) random initialization and a densification algorithm called Adaptive Bound-Expanding Split (ABE-Split) algorithm to enhance the robustness of 3DGS. However, SLV random initialization is incompatible with current other 3DGS-based models and cannot outperform 3DGS models with high-quality initializations.

\subsection{Frequency in 3D Reconstruction}

Frequency analysis has gained increasing attention in 3D studies with growing numbers of frequency-based methods than before. 
Zhang~\cite{zhang2023frequency} discovered that frequencies of 3D representations are region-dependent and refined texture details are main distributed on high frequencies according to , and they propose an adaptive frequency modulation module to learn local texture frequencies. 
NeRF~\cite{mildenhall2020nerf} utilizes a random Fourier feature mapping to help MLPs effectively learn high frequencies in low-dimension spaces~\cite{tancik2020fourier}.Barron~\cite{barron2022mip,yu2024mip} start from a frequency perspective and design anti-aliasing methods to achieve impressive performance on NeRF and 3DGS respectively. 
Zhang~\cite{zhang2024fregs} designs a frequency-related loss to solve the over-reconstruction issue.

\section{Preliminaries}

\subsection{Nyquist-Shannon Sampling Theorem}
\label{sec:nsst}
Nyquist-Shannon sampling theorem is a fundamental principle in signal processing that establishes the relationship between the frequency range of a signal and sampling rate required to avoid aliasing. The theorem states that to reconstruct a band-limited signal without aliasing, the sampling rate must be at least twice the bandwidth (the highest frequency) \(\nu\) of the signal.

Additionally, Yu~\cite{yu2024mip} developed a method to estimate sampling rates of Gaussians under the multi-view setting: For \(i^{th}\) Gaussian \(\mathrm p_i\), given N cameras, corresponding focal length \(f\) and depth \(d\), the  maximal sampling rate is defined as:
\begin{equation}
\nu_{i} \propto \max\limits_{k=1,\cdots,N}\left(\mathbf{1}_k(\mathrm p_i)\cdot\frac{f_k}{d_k}\right),
\end{equation}
and the sampling interval \(T\) is the inverse of sampling rate:
\begin{equation}
T = \frac{1}{\nu},
\end{equation}
where \(\mathbf{1}_k\) is an indicator denoting whether a Gaussian is visible from \(k^{th}\) camera. We adopt this method to estimate the sampling rate for each Gaussian in our subsequent analysis.

% \subsection{3D Gaussian Splatting and 3DGS-based editing}
\subsection{3D Gaussian Splatting}
\label{sec:3dgsfreq}
3DGS is a powerful explicit 3D representation method using the differential rendering algorithm. It employs Gaussian ellipsoids to represent 3D objects or scenes, which can be expressed using Eq.(\ref{eq:gauss}). Each Gaussian has SH coefficients (for representing color) and opacities (for \(\alpha\)-blending) apart from the center location \(p\) and the covariance matrix \(\Sigma=RSS^TR^T\), \(R\) represents the rotation matrix, and \(S\) represents the scaling matrix. \(R\) and \(S\) are represented by a quaternion \(q\) and a 3D vector \(s\) respectively.

Each 3D Gaussian primitive can be expressed as 
\begin{equation}
\mathcal G(x)=\frac{1}{(2\pi)^{\frac32}|\Sigma|^{\frac12}}\exp\left(-\frac12(x-p)^T\Sigma^{-1}(x-p)\right),
\label{eq:gauss}
\end{equation}
which can be represented as a weighted sum of different frequencies:
\begin{equation}
\mathcal G(x) = \frac{1}{(2\pi)^3}\int_{\mathbb{R}^n}\mathcal F(\mathcal G(x),\omega)e^{i\omega^T x}\d\omega,
\end{equation}
where 
\(
\mathcal F(\mathcal G(x),\omega)=\exp\left(-i\omega^Tp-\frac{\omega^T\Sigma\omega}{2}\right).
\)

Given an angular frequency \(\omega\), the corresponding frequency weight is
\(
|\mathcal F(\mathcal G(x),\omega)|=\exp\left(-\frac{\omega^T\Sigma\omega}{2}\right)
\),
which decays as the norm of the angular frequency \(\|\omega\|\) increases due to the positive-definiteness property of \(\Sigma\). 

\begin{figure}[htbp]
  \centering
    \includegraphics[width=\linewidth]{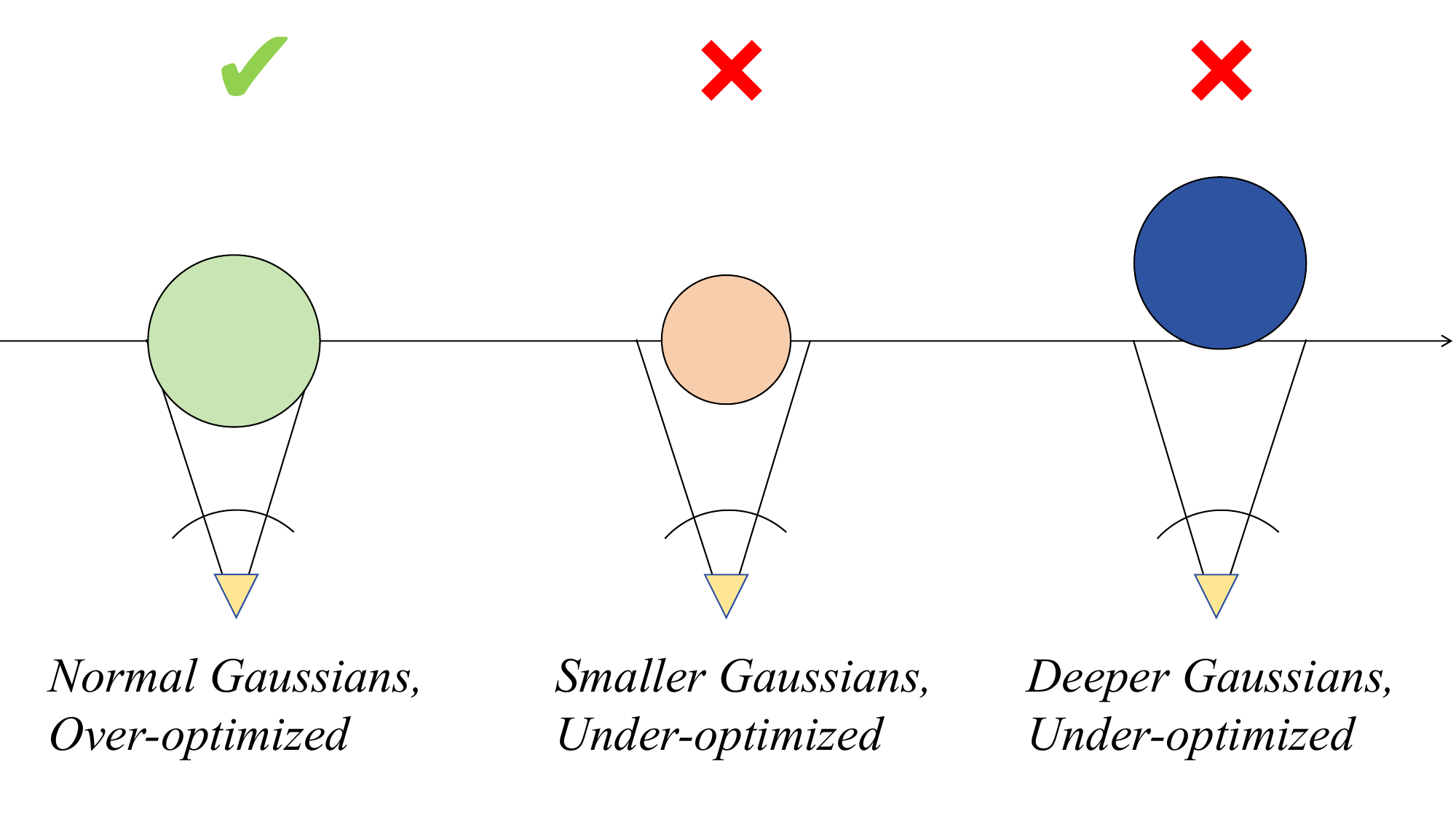}
  \caption{An demonstration diagram about optimizing Gaussians. According to the Nyquist sampling theorem, normal Gaussians bigger than corresponding sampling intervals are over-optimized, whereas smaller or deeper Gaussians whose scales cannot surpass the sampling intervals are under-optimized.}
  \label{fig:demo}
\end{figure}

\(R\) is an orthogonal matrix, so \(S\) determines the frequency components of a Gaussian primitive.
To be specific, for a Gaussian primitive, a larger scaling matrix \(S\) leads to a narrower bandwidth in the frequency domain. Hence, a larger Gaussian contains relatively more low-frequency information and vice versa.

3DGS is sensitive to noise in the initialization and noise prevents it from achieving highly qualified results: It usually achieves high performance with accurate high-quality initialization priors (SfM initialization~\cite{trackgs}, LiDAR data~\cite{Jiang2025icra}, etc.) Experiments in Fig.~\ref{fig:noisyexp} also show that 3DGS cannot help Gaussians learn low frequencies better when the model is initialized with noise.

\subsection{Spectral Bias}
\label{sec:spectral}

Rahaman~\cite{rahaman2019spectral} found the spectral bias in training neural networks: Neural networks usually learn low-frequency information first, and then gradually capture high-frequency information.
Spectral bias is usually related to the generalization and performance of neural networks~\cite{fridovich2022spectral,tancik2020fourier,wang2020high}, so it is an important property of networks.

Interestingly, we find similar phenomenon in training 3DGS-based models (Fig.~\ref{fig:numofgauss} (c)). We also find the similar phenomenon in training 2DGS, more experiments are described in Appendix. B. According to the analysis in Sec.~\ref{sec:3dgsfreq}, smaller Gaussians contain more high-frequency information and less low-frequency information. However, it is unknown whether this phenomenon always exists in training various 3DGS-based models on various scenes. More comprehensive experiments are needed to study the behavior of 3DGS-based models in the frequency domain during the training process.

\section{Method}
\label{sec:meth}
In this section, we analyze floating artifacts in 3DGS from a frequency perspective. 
We find that floating artifacts arise because Gaussians are over-shrunk and difficult to optimize. These over-shrunk Gaussians have not learned accurate low-frequency information well before diving into fine-grained high-frequency information.
After that, we propose \textit{Eliminating-Floating-Artifacts} Gaussian Splatting (EFA-GS) to mitigate these artifacts.

\subsection{Impacts of Low-quality Initialization}
\label{sec:expanal}
Before analyzing floating artifacts, we make the following hypothesis.

\noindent
\textbf{Hypothesis 1.} \textit{Low quality initialization can be viewed as accurate initialization with noise.}

The hypothesis is likely correct for most situations, so we use noisy initialization to represent low-quality initialization in the following experiments and analyses.

To investigate the cause of floating artifacts when the model lacks high-quality SfM initializations, we perform experiments on ship from the NeRF Synthetic dataset~\cite{mildenhall2020nerf} and kitchen from the Mip-NeRF 360 dataset~\cite{barron2022mip}. Specifically, we introduced noise to the SfM initialization of Gaussian scales and coordinates before training. Details and more experimental results are provided in the Appendix. A. Experiments indicate that
\begin{itemize}
    \item [1.] As illustrated in Fig.~\ref{fig:numofgauss}, 3DGS-based models (Mip-splatting) sometimes have a frequency bias:  they rapidly shrink at first (about 5,000 iterations). In addition, corrupted initialization is related to excessive shrinkage of most Gaussians, which means that floating artifacts are probably related to a lack of accurate low frequencies.
    \item [2.] As shown in Fig.~\ref{fig:noisyexp} and Tab.~\ref{tab:noisyexp}, floating artifacts emerged with noisy initialization, significantly degrading the final rendering quality. Moreover, these artifacts harm more severely to the rendering quality of testing views than training views, thereby expanding the gap between training and testing views.
\end{itemize}

Based on these findings, we provide our theoretical analysis in the next subsection.

\subsection{Theoretical Analysis of Floating Artifacts}
\label{sec:theanal}

According to the sampling theorem, the sampling rate must be higher than the Nyquist sampling rate to accurately reconstruct a signal. 
Therefore, there are two key factors that determine whether a Gaussian can be sufficiently optimized: scale and coordinate. 
In Multi-View Synthesis (MVS) tasks, scale determines the frequency information that a Gaussian can receive, and coordinate determines the depth corresponding to different viewpoints, which is directly related to sampling rates (intervals), as mentioned in Sec.~\ref{sec:nsst}. 
If the scale of a Gaussian surpasses the corresponding sampling interval, the Gaussian can probably be sufficiently optimized, such a Gaussian is probably an over-optimized Gaussian. 
Otherwise, the Gaussian is probably an under-optimized Gaussian, as illustrated in Fig.~\ref{fig:demo}. 

According to results in Fig.~\ref{fig:numofgauss} and ~\ref{fig:noisyexp} (a) and (b), noises would force more Gaussians to become over-shrunk Gaussians and floating artifacts can be observed at the same time.
This relationship shows that inaccurate low-quality initializations may cause the existence of floating artifacts. However, the original initializations of ship~\cite{mildenhall2020nerf} are also a bunch of random Gaussian points, as mentioned in the Appendix. A. Training Mip-splatting does not introduce massive floating artifacts until noises are injected. Hence, at least inaccurate initialization is not the sufficient condition resulting in floating artifacts.

Since floating artifacts are related to the over-shinkage of Gaussians, we propose a proposition:

\noindent
\textbf{Proposition 1.} \textit{In low-quality initialization scenarios, inaccurate under-optimized Gaussians would be more likely to become floating artifacts than inaccurate over-optimized ones in the optimization process.}

Noise is inherently random, so it would result in more inaccurate over-optimized Gaussians and inaccurate under-optimized ones for each camera simultaneously.
However, according to the proposition, for training cameras inaccurate over-optimized Gaussians can be sufficiently corrected during training because their scales are larger than sampling intervals. 
No such correction occurs for testing cameras because they do not participate in the training process. 
Hence, if our proposition holds, noise should lead to more performance degradation for testing views than training views. Otherwise, noise would cause equal degradation to the training and testing views.

We conduct experiments in Fig.~\ref{fig:noisyexp} (a), (b) and record average PSNR score from training and testing views respectively in Tab.~\ref{tab:noisyexp} to observe the performance degradation caused by noisy initialization. The quantitative results support our inferences: noise in the initialization actually results in more damage for rendered images of testing views than ones of training views.

We also conduct sparse-view experiments to explore the connection between under-optimized Gaussians and floating artifacts. Experiments in Appendix. A indicate that Under-optimized Gaussians are more sensitive to noise and easily become artifacts.

\paragraph*{\textbf{Why Mip-splatting failed?}} Since floating artifacts are related to inaccurate coordinates and scales, there are two ways to solve this issue: 1. Depth regularization. Previous researches~\cite{Huang2DGS2024} already developed depth-related regularization losses to enhance the performance of 3DGS; 2. Low-pass filter for Gaussians. It is natural to design a low-pass filter to force Gaussians to learn accurate low-frequency information first. Previous work like Mip-splatting~\cite{yu2024mip} actually did the same thing. However, experiments in Fig.~\ref{fig:noisyexp} (b) show that Mip-splatting failed in removing artifacts. Here we provide a possible explanation: Experimental results in Fig.~\ref{fig:numofgauss} (c) illustrates that Gaussians are rapidly shrunk at first and noise exacerbates this phenomenon. Many Gaussians are suddenly over-shrunk before they extract accurate frequency information.  Therefore, a fixed low-pass filter is possibly not enough to help these Gaussians learn accurate low-frequencies. More importantly, over-optimized Gaussians and under-optimized ones should be treated differently, whereas Mip-splatting applies the same low-pass filter to Gaussians having the same depths. So it is necessary to select under-optimized Gaussians and directly manipulate their scales during the optimization process.

\paragraph*{\textbf{Limitations of Depth-based regularization}} Depth and scale are 2 important factors in optimizing Gaussians. However, depth-based regularization usually has one drawback: depths of Gaussians are relatively fixed. As shown in Fig.~\ref{fig:numofgauss} (c), the scales of Gaussians rapidly shrink, and noise in the initialization exacerbates this phenomenon. So depths sometimes offer limited help to under-optimized Gaussians. Results in Fig.~\ref{fig:noisyexp} (b) and (c) are consistent with our conclusions: Gaussians should have relatively close depths for similar camera spaces, whereas floating artifacts sometimes cannot be observed from the training view even if the training view is close to the testing view. To explore the limitation of deph-regularization method, we conduct experiments on the depth-related regularization of 2DGS~\cite{Huang2DGS2024}, more details can be found in Appendix. B.

% According to Nyquist-Shannon Sampling algorithm, it is possible to accurately reconstruct frequency components up to \(\frac{\nu}{2}\) with given sampling rate \(\nu\). 
% Consequently a Gaussian smaller than \(2T\) may not be well reconstructed. 
% On the other hand, a smaller Gaussian has a smaller receptive field in the corresponding 2D pixel space, making it more difficult for a smaller Gaussian to learn those global low-frequency information well, which is usually fit first in training other models like neural networks~\cite{rahaman2019spectral}.

\begin{algorithm}[!t]
    \textsl{}
    \caption{LFCF algorithm}
    \label{alg:lfcf}
    \begin{algorithmic}[1]
        \Require previous gradients \(PGrad(\cdot)\), current gradients \(Grad(\cdot)\), enlarging factor \(s(\cdot)\), processing threshold \(\tau\), opacities \(\alpha(\cdot)\), opacity threshold \(\epsilon\), splitting threshold \(\eta(\cdot)\)
        % \Ensure generated sample \(\mathrm x\)
        % \State \(i\leftarrow0\)
        \For{\(i^{th}\) Gaussian in all Gaussians}
            \If{\(Grad(i)>\tau\)}
                \If{\(Grad(i)>PGrad(i)\)}
                    \State ExpandGaussian(\(i\),\(s(i)\))
                \Else
                    \State ShrinkGaussian(\(i\))
                    \State SplitGaussian(\(i\))
                    % \State Sample \(p\in U[0,1]\)
                    % \If{\(p>\eta(i)\)}
                    %     \State SplitGaussian(\(i\))
                    % \EndIf
                \EndIf
            \EndIf
            \State \(PGrad(i)\leftarrow Grad(i)\)
        \EndFor
        % \State \(i\leftarrow0\)
        \For{\(i^{th}\) Gaussian in all Gaussians}
            \If{\(\alpha(i)<\epsilon\)}
                \State RemoveGaussian(\(i\))
            \EndIf
        \EndFor
    \end{algorithmic}
\end{algorithm}

\subsection{EFA-GS}
\label{sec:efags}

Although some depth-based regularization methods have been proposed, they usually have some limitations in experiments (erosion of delicate details, etc.) So we propose our EFA-GS, an enhanced 3DGS in order to effectively mitigate floating artifacts. EFA-GS consists of the Low-Frequency-Come-First (LFCF) algorithm and some strategies.

\paragraph*{\textbf{LFCF algorithm}} The core idea of LFCF algorithm is to \textbf{selectively expand Gaussians} during training. According to Fig.~\ref{fig:numofgauss}, Gaussians gradually shrink in training and smaller Gaussians receive fewer rays and are projected to fewer pixels during the rendering process, which means their gradients are usually smaller than those of larger Gaussians. Therefore, gradient ascent is an anomaly caused by other factors (especially floating artifacts). Proposition 1 in Sec.~\ref{sec:theanal} indicates over-optimized Gaussians are less likely to become artifacts, so if a Gaussian has a larger gradient than before, we consider it to be an under-optimized Gaussian becoming artifacts, then the LFCF algorithm expands the Gaussian to force it to fit low frequencies first before diving into complicated high-frequency details as shown in Alg.~\ref{alg:lfcf};
if a Gaussian has a smaller gradient than before, we consider it to be an over-optimized Gaussian, and the LFCF algorithm shrinks and splits it. Both the expanding and shrinking operations are implemented as \(s^{\prime}=s\cdot c\), where \(c\in\mathbb{R}^3\) controls expansion (\(c>1\)) or shrinkage (\(c<1\)).
The experimental results in Fig.~\ref{fig:noisyexp} and Tab.~\ref{tab:noisyexp} show the effectiveness of the LFCF algorithm.

\begin{figure}[htbp]
  \centering
  \begin{subfigure}{0.32\linewidth}
    \includegraphics[width=\linewidth]{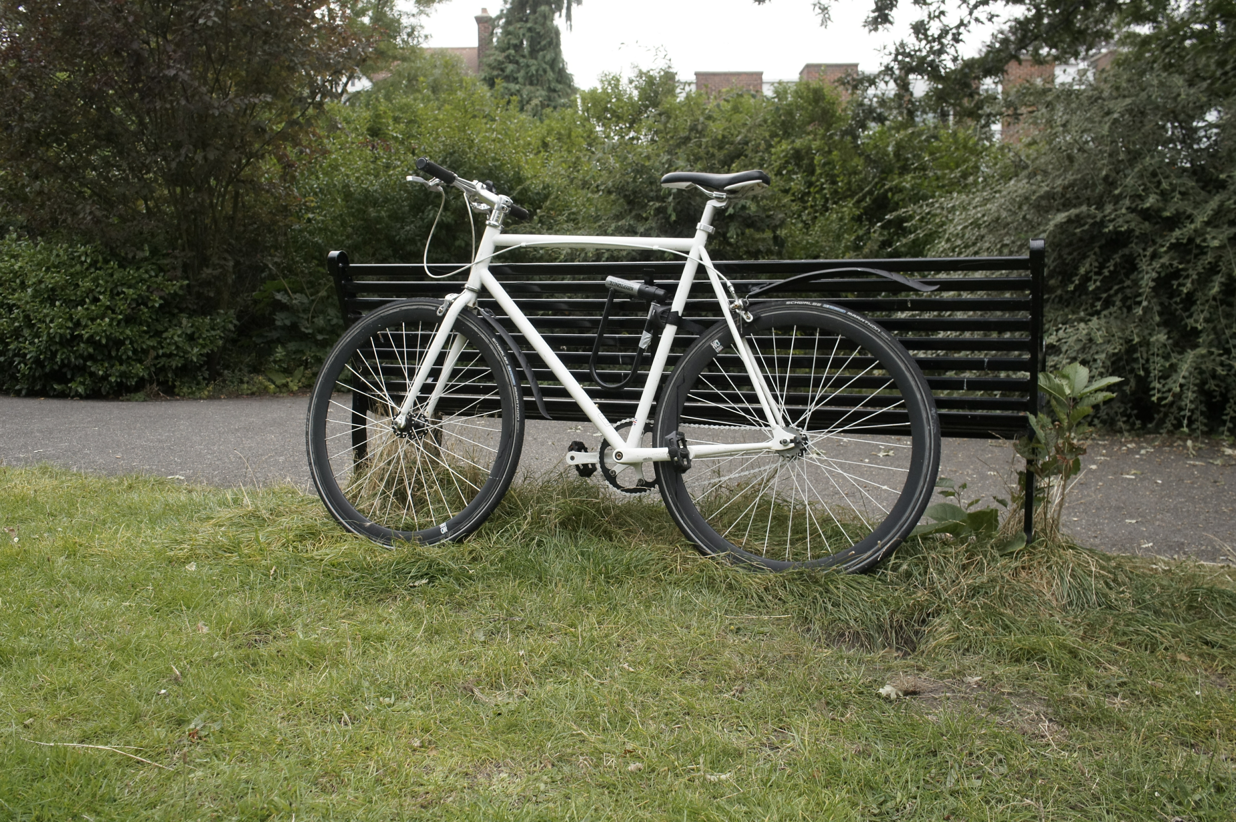}
    \caption{Ground truth.}
  \end{subfigure}
  \begin{subfigure}{0.32\linewidth}
    \includegraphics[width=\linewidth]{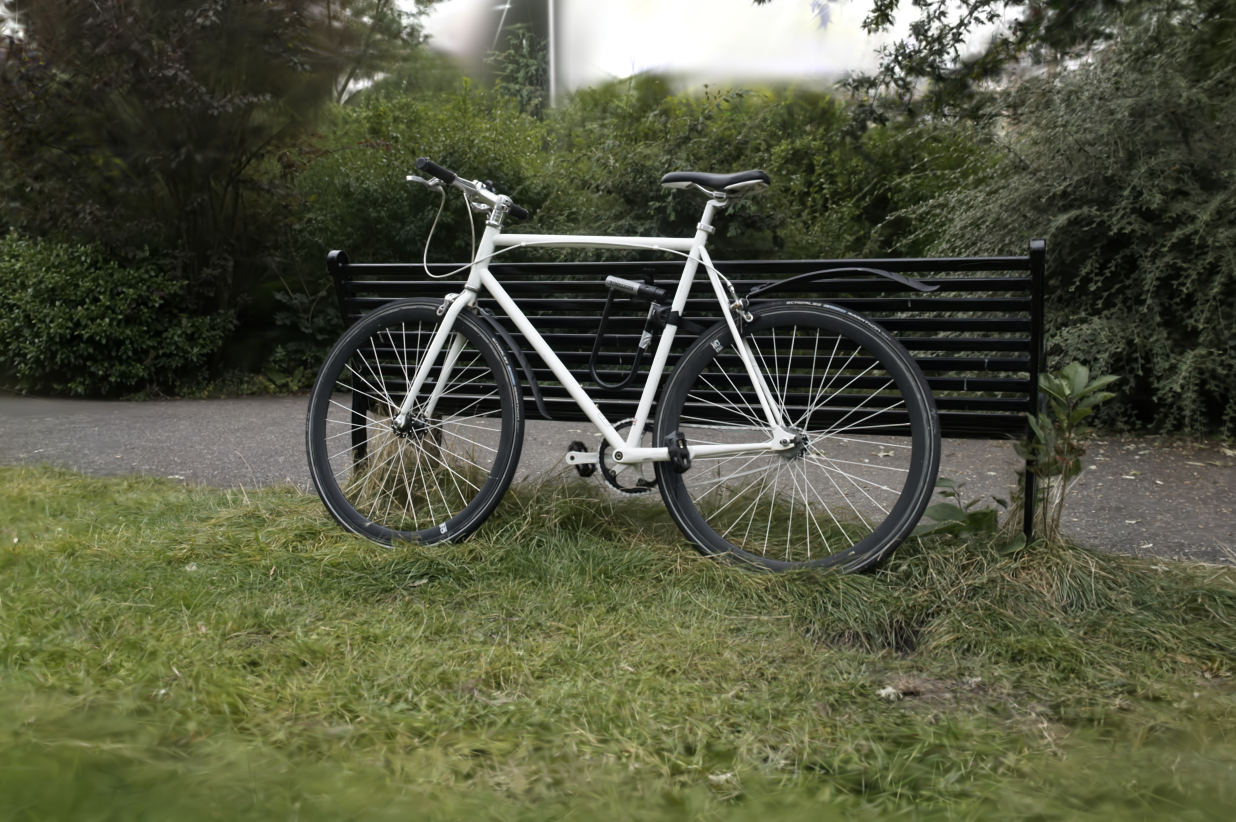}
    \caption{LFCF.}
  \end{subfigure}
  \begin{subfigure}{0.32\linewidth}
    \includegraphics[width=\linewidth]{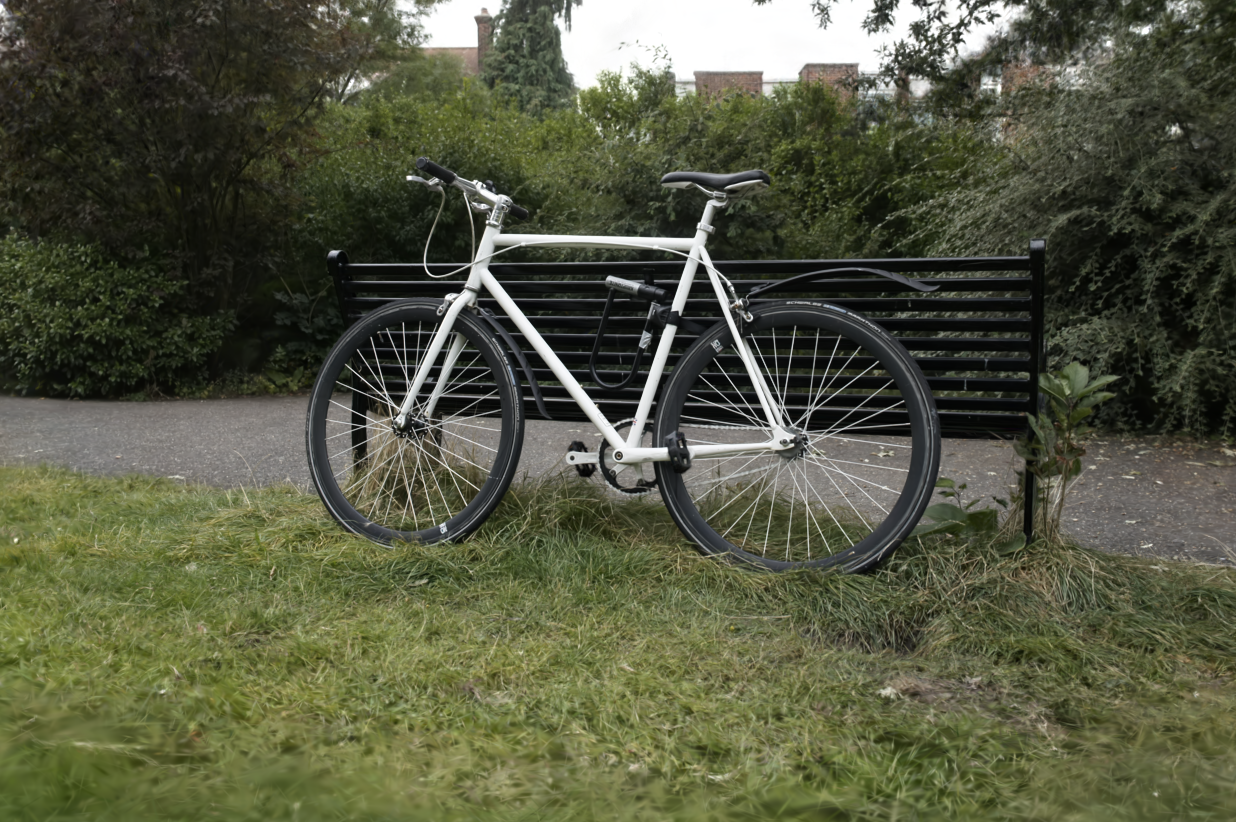}
    \caption{EFA-GS.}
  \end{subfigure}
  \caption{Reconstruction results of bicycle model(outdoors). The results shows that erosion of the buildings would arise if the strategies were not used.}
  \label{fig:simple}
\end{figure}

While the LFCF algorithm works well in simple 3D object datasets, simply setting hyperparameters such as the enlarging factor \(c(\cdot)\) leads to erosion of the details in the distance when reconstructing complicated 3D scenes (Fig.~\ref{fig:simple}). This tradeoff may restrict the application of the LFCF algorithm. So, it is necessary to design several strategies to mitigate this problem. Here we propose depth/scale-based strategies to solve this issue. To further achieve fine-grained reconstruction results, we also design several other strategies and they are described in Appendix C. All strategies are independent from each other.

\paragraph*{\textbf{Depth-based Strategy}} We observe that deeper Gaussians have lower sampling rates and are more difficult to optimize, so we assign deeper Gaussians lower enlarging factors. To be specific, we first calculate the sampling rates of Gaussians. For simplicity, we adopt the definition of sampling rate in Mip-splatting~\cite{yu2024mip}.
Then we project them into [0,1] as linear interpolation factors \(\theta(\cdot)\).
After that we introduce 2 hyperparameters (\(c_{max}\) and \(c_{min}\)) for linear interpolation. For \(i^{th}\) Gaussian, the enlarging factor \(c(i)\) is calculated using following formula:
\begin{equation}
c(i) = \theta(i)\cdot c_{max} + \left(1-\theta(i)\right)\cdot c_{min}.
\end{equation}

\(c_{min}\) is usually set to 1.

\paragraph*{\textbf{Scale-based Strategy}} The strategy treats scaling factors of different axes differently because sampling intervals of different axes are different. To preserve the Gaussian volume, we design a volume-preserving setting: \(\prod_ic_i=1\). With this setting, LFCF expands the shortest axis and shrinks the longest one.

We also design 3 other strategies, and more technical details are provided in the Appendix. C.

\section{Experiments}

We first present technical details in our implementation, then release the performance of EFA-GS on various 3D tasks. We also perform ablation studies to EFA-GS. 
% \textbf{Due to the limited space, we provide massive experiments in Appendix: Additional ablation studies in Appendix. C; Speed experiments in Appendix. E; Qualitative comparisons in Appendix. F; Hyperparameter sensitivity analysis in Appendix. G.} Speed experiments demonstrate that EFA-GS efficiently eliminate artifacts.

\subsection{Details in Implementation and Comparisons}

\begin{table}[htbp]
    \caption{Reconstruction results on RWLQ dataset. Our EFA-GS achieves state-of-the-art performance among these 3DGS-based methods. EFA-GS(Mip) outperforms the baseline (Mip-splatting) by 1.68 (PSNR) and 2DGS by 0.35 (PSNR); EFA-GS(3DGS) outperforms the baseline (Vanilla 3DGS) by 0.99 (PSNR) and 2DGS by 0.70 (PSNR). The best score of \colorbox{CoralRed!50}{EFA-GS} and \colorbox{Apricot!90}{other methods} are \colorbox{CoralRed!50}{coralred} and \colorbox{Apricot!90}{apricot} respectively.}
    \centering
    \begin{tabular}{c|ccc}
        \toprule
                      & PSNR\(\uparrow\) & SSIM\(\uparrow\) & LPIPS\(\downarrow\) \\
        \midrule
        2DGS~\cite{Huang2DGS2024} & \cellcolor{Apricot!90}\textbf{28.00} & \cellcolor{Apricot!90}\textbf{0.95} & 0.16  \\
        GOF~\cite{yu2024gaussian} & 27.92 & 0.95 & \cellcolor{Apricot!90}\textbf{0.15}  \\
        eRank-GS~\cite{erankgs} & 23.16 & 0.91 & 0.18  \\
        \midrule
        Vanilla 3DGS~\cite{kerbl3Dgaussians} & 27.71 & 0.95 & 0.15  \\
        EFA-GS(3DGS) & \cellcolor{CoralRed!50}\textbf{28.70} & \cellcolor{CoralRed!50}\textbf{0.95} & \cellcolor{CoralRed!50}\textbf{0.14} \\
        \midrule
        Mip-splatting~\cite{yu2024mip} & 26.67 & 0.94 & 0.16  \\
        EFA-GS(Mip, default)  & 28.35 & 0.95 & 0.15  \\
        \bottomrule
    \end{tabular}
    \label{tab:resofrwlq}
\end{table}

For 3D reconstruction tasks, we set \(c_{max}\) to 1.5 and \(c_{min}\) to 1.

For quantitative comparisons, we choose PSNR, SSIM and LPIPS as evaluation metrics. We compare our EFA-GS with 3DGS~\cite{kerbl3Dgaussians}, Mip-splatting~\cite{yu2024mip}, 2DGS~\cite{Huang2DGS2024}, Gaussian Opacity Fields (GOF)~\cite{yu2024gaussian} and eRank-GS~\cite{erankgs}. We adopt 3DGS and Mip-splatting as baselines and implement EFA-GS on them. The default setting is EFA-GS(Mip) using Mip-splatting as baseline. We choose 2DGS and GOF because they use depth-based regularization method to enhance performance, and we choose eRank-GS because eRank-GS can eliminate needle-like artifacts. Qualitative comparisons are provided in Appendix. F for comprehensive comparisons.

% We execute official codes of each method on all datasets and report average experimental results except for eRank-GS on Mip-NeRF360 dataset. For eRank-GS on Mip-NeRF360, we just use the experimental results provided in the original paper.

\subsection{Datasets}

Situations in the real world are often more complicated than injecting noises into high-quality initializations. To better evaluate EFA-GS, we conduct experiments on Real-World-Low-Quality(RWLQ), Mip-NeRF360~\cite{barron2022mip} and TanksandTemples~\cite{Knapitsch2017} dataset.

To better evaluate our method on real-world low-quality scenarios, we present a dataset named the RWLQ dataset. We collect 4 realistic scenes using cellphone cameras: astronaut, pheasant, porcelain and tank. Due to several reasons(missing viewpoints, low-quality photos, etc.), 3DGS and Mip-splatting usually produce floating artifacts on RWLQ dataset. 

TanksandTemples is a real-world complex dataset containing many scenes. Unlike previous work~\cite{Huang2DGS2024}, we select more challenging scenes in TanksandTemples dataset to adequately evaluate our EFA-GS. To be specific, we choose 6 advanced complex scenes: Auditorium, Ballroom, Courtroom, Museum, Palace and Temple. Moreover, we find a method that can increase the numbers of initialization Gaussian points and lower down the final performance without injecting noises. More details are described in the Appendix. D. We also perform experiments using low-quality initialization on the TanksandTemples dataset.

For all datasets we use Colmap~\cite{Schonberger_2016_CVPR} to generate a sparse point cloud first, and then use it as initialization priors for all methods.

\subsection{Quantitative Comparisons}
\paragraph*{Comparisons on RWLQ Dataset}
As illustrated in Tab.~\ref{tab:resofrwlq}, EFA-GS surpasses all other methods and achieves state-of-the-art performance. EFA-GS(Mip) significantly improves performance compared to its baseline (Mip-splatting) by 1.68(PSNR) and surpasses other depth-based regularization methods like 2DGS(0.35), GOF(0.43), demonstrating the effectiveness of EFA-GS. Qualitative comparison are shown in Appendix. F.

\begin{table}
    \caption{Reconstruction results on Mip-NeRF360 dataset. Mip-NeRF360 is a high-quality dataset. 3DGS (or Mip-splatting) does not produce many floating artifacts that can be easily observed. However, our EFA-GS(Mip) still slightly outperforms baseline (Mip-splatting). The results shows that the tradeoff mentioned in Sec.~\ref{sec:efags} has been successfully handled, and EFA-GS does not produce blurrier results when there are only a few floating artifacts in the reconstructed scene.}
    \centering
    \begin{tabular}{c|ccc}
        \toprule
                      & PSNR\(\uparrow\) & SSIM\(\uparrow\) & LPIPS\(\downarrow\) \\
        \midrule
        2DGS          & 27.00 & 0.81 & 0.24  \\
        GOF           & 27.33 & 0.82 & 0.20  \\
        eRank-GS      & 27.69 & 0.84 & 0.20  \\
        \midrule
        Vanilla 3DGS  & 27.58 & 0.82 & 0.21  \\
        EFA-GS(3DGS)  & 27.52 & 0.82 & 0.21  \\
        \midrule
        Mip-splatting & \cellcolor{Apricot!90}\textbf{27.92} & \cellcolor{Apricot!90}\textbf{0.84} & \cellcolor{Apricot!90}\textbf{0.18}  \\
        EFA-GS(Mip, default)  & \cellcolor{CoralRed!50}\textbf{27.94} & \cellcolor{CoralRed!50}\textbf{0.84} & \cellcolor{CoralRed!50}\textbf{0.18}  \\
        \bottomrule
    \end{tabular}
    \label{tab:resofmip}
\end{table}

\paragraph*{Comparisons on Mip-NeRF360 Dataset}

\begin{table}
    \caption{Reconstruction results on TanksandTemples dataset. EFA-GS(Mip) outperforms baseline (Mip-splatting) by 0.68 (PSNR) on normal initialization and 0.92 on low-quality initialization; EFA-GS(3DGS) outperforms baseline (Vanilla 3DGS) by 0.18 (PSNR) on normal initialization and 0.34 on low-quality initialization.}
    \resizebox{0.48\textwidth}{!}{
    \begin{tabular}{c|ccc|ccc}
        \toprule
                      & \multicolumn{3}{c}{Overall} & \multicolumn{3}{c}{Low-Quality Init} \\
                      & PSNR\(\uparrow\) & SSIM\(\uparrow\) & LPIPS\(\downarrow\) & PSNR\(\uparrow\) & SSIM\(\uparrow\) & LPIPS\(\downarrow\)\\
        \midrule
        2DGS          & 21.17    & 0.78   & 0.32    & 18.98    & \cellcolor{Apricot!90}\textbf{0.72}   & 0.41    \\
        GOF           & 19.80    & 0.77   & 0.30    & 17.84    & 0.67   & 0.39    \\
        eRank-GS      & 18.43    & 0.72   & 0.37    & 16.37    & 0.62   & 0.45    \\
        \midrule
        Vanilla 3DGS  & \cellcolor{Apricot!90}\textbf{21.51}    & \cellcolor{Apricot!90}\textbf{0.79}   & \cellcolor{Apricot!90}\textbf{0.28}    & \cellcolor{Apricot!90}\textbf{19.11}    & 0.69   & \cellcolor{Apricot!90}\textbf{0.37}    \\
        EFA-GS(3DGS)  & \cellcolor{CoralRed!50}\textbf{21.69}    & \cellcolor{CoralRed!50}\textbf{0.80}   & \cellcolor{CoralRed!50}\textbf{0.28}    & \cellcolor{CoralRed!50}\textbf{19.45}    & \cellcolor{CoralRed!50}\textbf{0.69}   & \cellcolor{CoralRed!50}\textbf{0.36}    \\
        \midrule
        Mip-splatting & 20.63    & 0.78   & 0.29    & 18.15    & 0.67   & 0.39    \\
        EFA-GS(Mip, default)  & 21.31    & 0.79   & 0.28    & 19.07    & 0.69   & 0.37    \\
        \bottomrule
    \end{tabular}
    }
    \label{tab:resoftat}
\end{table}

Mip-NeRF360 is a real-world high-quality dataset. Due to high-quality initialization and sufficient training views, 3DGS results on high-quality datasets (Mip-NeRF360) usually contain only a few artifacts, which have limited effects on the results. Instead, it is more necessary for EFA-GS not to enhance performance but to carefully handle the trade-off mentioned in Sec.~\ref{sec:efags}. As shown in Tab.~\ref{tab:resofmip}, EFA-GS successfully handles the tradeoff: EFA-GS(Mip) slightly outperforms Mip-splatting and other methods. Qualitative comparison are shown in Appendix. F.

\begin{table}[htbp]
    \caption{Ablation studies on Mip-NeRF360 dataset. Ablation results reveal that each strategy of EFA-GS is useful and effective.}
    \resizebox{0.48\textwidth}{!}{
    \begin{tabular}{c|cc|ccc}
        \toprule
                      & Depth & Scale & PSNR\(\uparrow\) & SSIM\(\uparrow\) & LPIPS\(\downarrow\) \\
        \midrule
        EFA-GS                   & \(\checkmark\) & \(\checkmark\) & 27.94 & 0.84 & 0.18  \\
        EFA-GS(w/o depth)        &   & \(\checkmark\) & 27.83 & 0.83 & 0.18  \\
        EFA-GS(w/o scale)        & \(\checkmark\) &   & 27.91 & 0.83 & 0.18  \\
        EFA-GS(w/o scale\&depth) &   &   & 27.37 & 0.83 & 0.19  \\
        \bottomrule
    \end{tabular}
    }
    \label{tab:ablation}
\end{table}

\paragraph*{Comparisons on TanksandTemples Dataset}

\begin{figure}[htbp]
  \centering
  \begin{subfigure}{0.32\linewidth}
    \includegraphics[width=\linewidth]{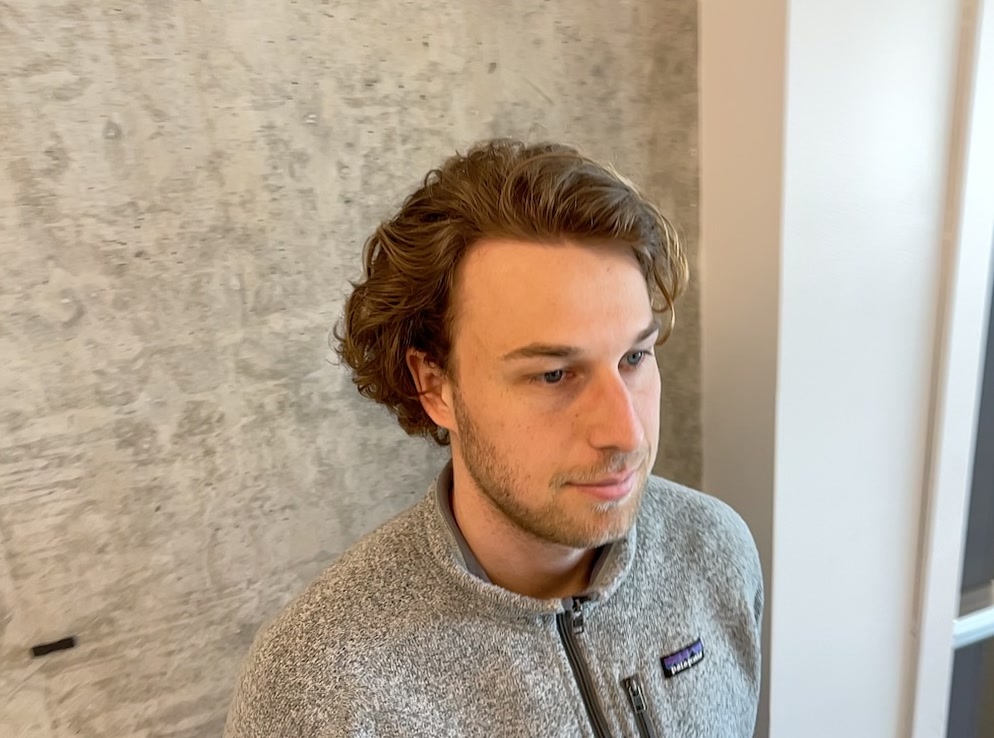}
    \includegraphics[width=\linewidth]{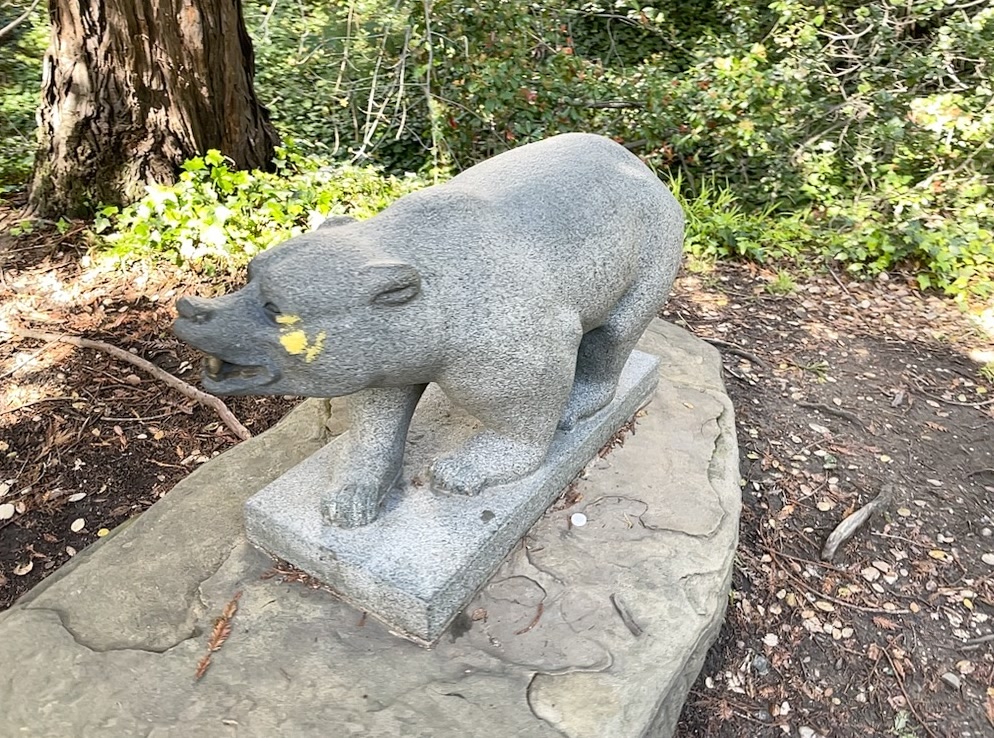}
    % \vspace{5pt}
    % \includegraphics[width=\linewidth]{picture/gt_face_2.jpg}
    \caption{Original View.}
  \end{subfigure}
  % \hfill
  \begin{subfigure}{0.32\linewidth}
    \includegraphics[width=\linewidth]{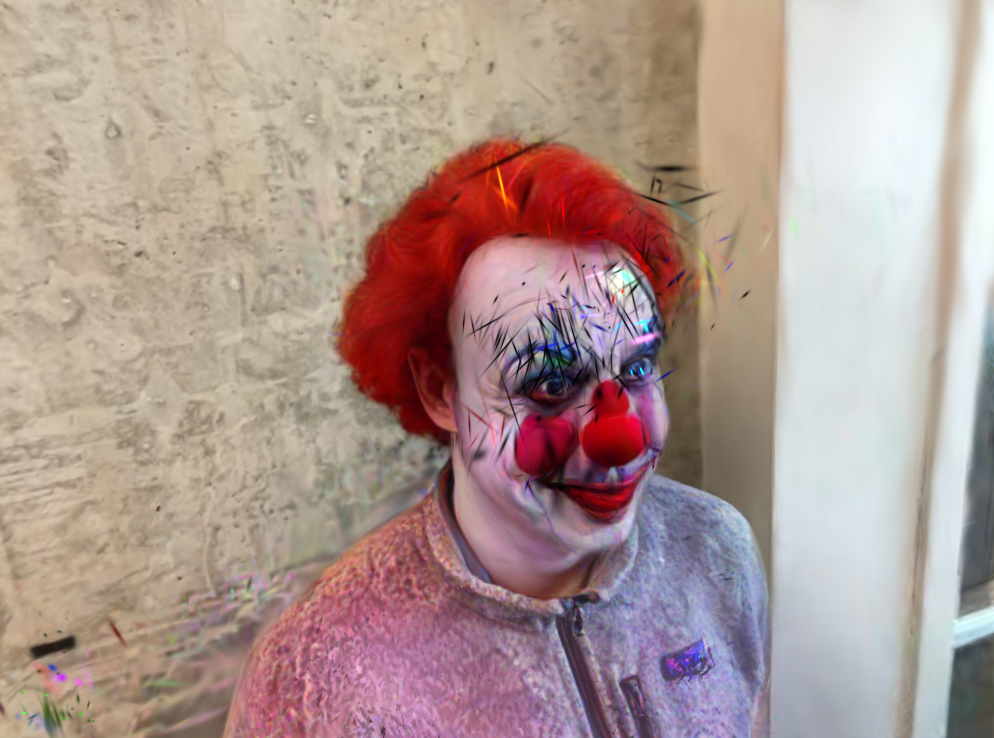}
    \includegraphics[width=\linewidth]{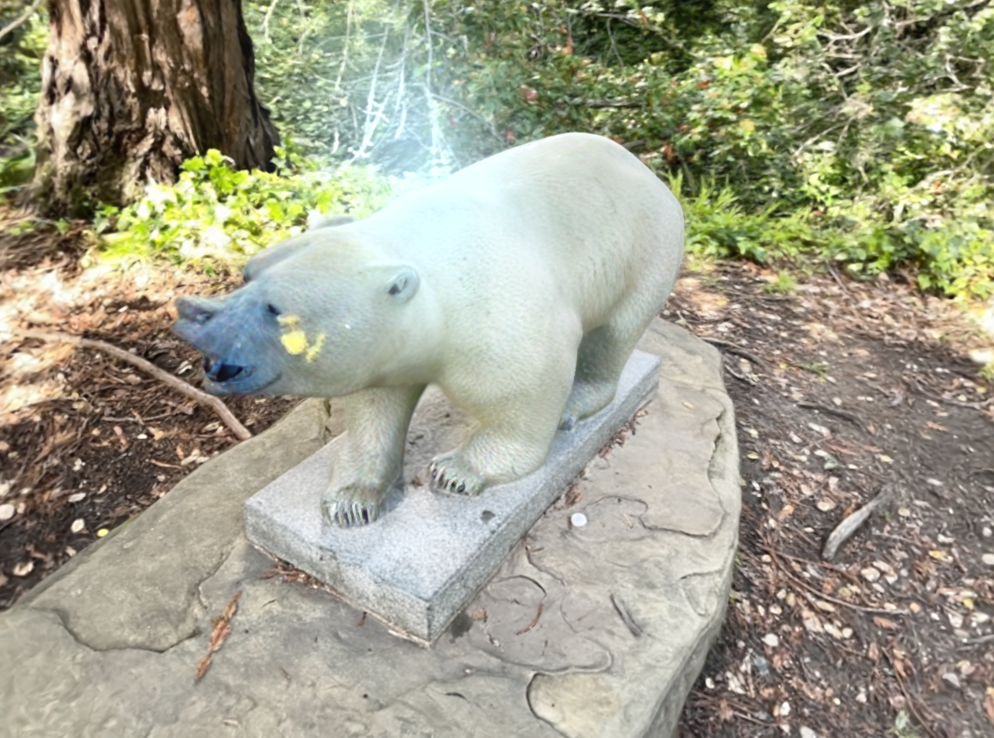}
    % \vspace{5pt}
    % \includegraphics[width=\linewidth]{picture/original_hulk_2.png}
    \caption{GE~\cite{chen2024gaussianeditor}.}
  \end{subfigure}
  \begin{subfigure}{0.32\linewidth}\centering
   \includegraphics[width=\linewidth]{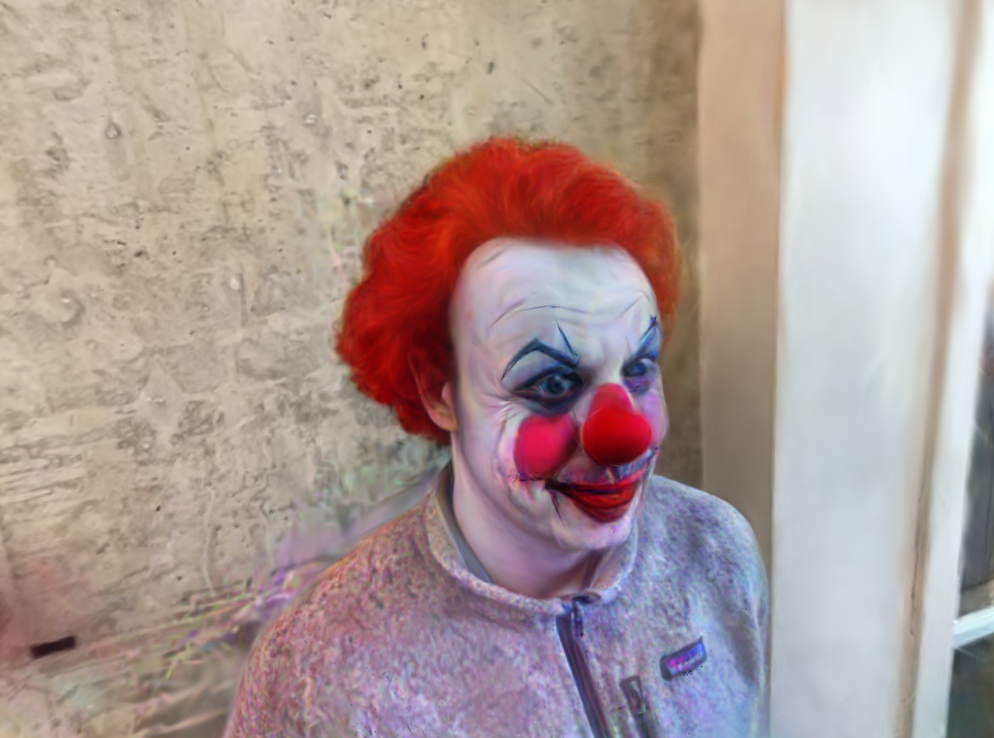}
   \includegraphics[width=\linewidth]{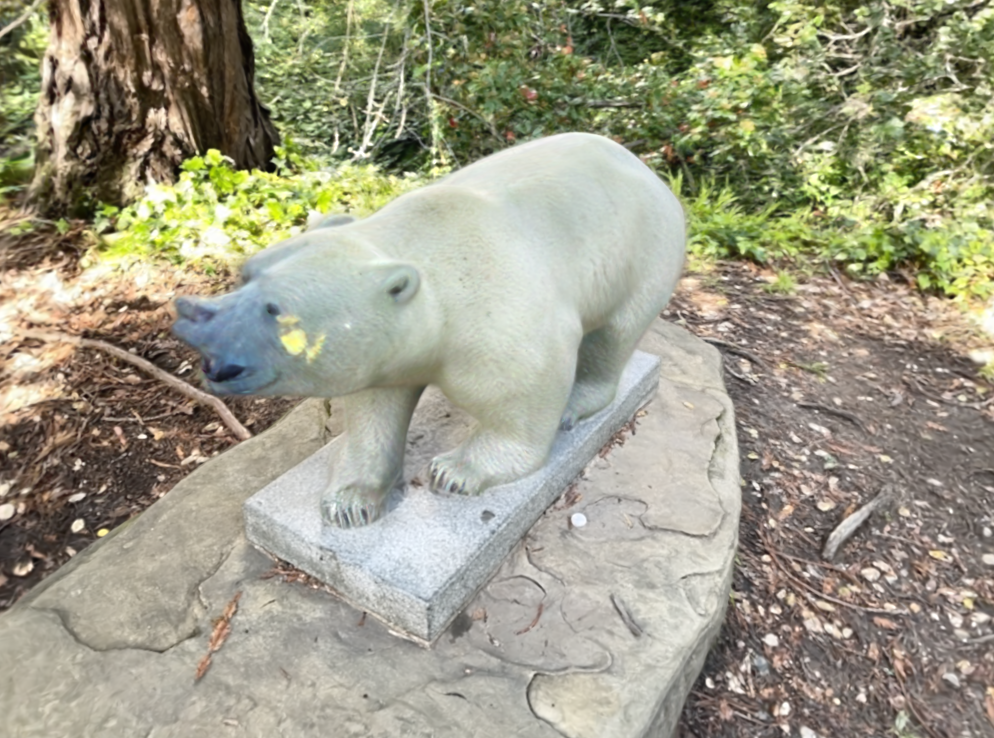}
    % \vspace{5pt}
    % \includegraphics[width=\linewidth]{picture/improved_hulk_2.png}
    % \caption{(log2)numbers of splitted Gaussians.}
    \caption{GE+EFA-GS.}
  \end{subfigure}
  \caption{Editing results of face and bear dataset~\cite{Haque_2023_ICCV}. \textbf{The first row} shows the editing results of editing command \textit{``Turn him into a clown"}, and \textbf{the second row} shows the editing results of editing command \textit{``Turn it into polar bear"}. Original GaussianEditor sometimes produces artifacts in the editing process, and EFA-GS effectively mitigates this issue. }
  \label{fig:edit}
\end{figure}
As shown in Tab.~\ref{tab:resoftat}, EFA-GS(3DGS) improves performance compared to 3DGS by 0.18(PSNR), 2DGS(0.52), GOF(1.89). Moreover, Low-quality results show that EFA-GS(3DGS) improves performance compared to 3DGS by 0.34(PSNR), 2DGS(0.47), GOF(1.61). EFA-GS(Mip) also achieves significant performance improvement on Mip-splatting. We also implement EFA-GS on GOF/erank-GS and perform experiments on this dataset. Experiments are provided in Appendix. E. Qualitative comparison are shown in Appendix. F.
% Although 2DGS successfully removes artifacts, it produces blurrier results in certain regions.

\subsection{Ablation Studies}

We perform comprehensive ablation studies to evaluate the contribution of the two key strategies: the depth-based strategy and the scale-based strategy. Experiments are summarized in Table~\ref{tab:ablation}. Initially, we ablate the depth-based strategy and observe a significant performance degradation. Subsequently, we remove the scale-based strategy independently, which yields only marginal performance differences. To further investigate the role of the scale-based strategy, we conduct an ablation by disabling both strategies simultaneously. By comparing these results, we demonstrate that both strategies collectively contribute to the performance enhancement of EFA-GS. Other strategies are described in Appendix. C and their ablation studies are provided in Appendix. C.

\subsection{Experiments on 3D Editing Tasks}

We also perform experiments on other multi-modal tasks like text-driven 3D Editing. We train a Mip-splatting model on the face and bear dataset~\cite{Haque_2023_ICCV} and then use GaussianEditor~\cite{chen2024gaussianeditor} to do 3D editing tasks. As illustrated in Fig.~\ref{fig:edit}, EFA-GS effectively mitigates floating artifacts.

\section{Conclusions}

We present EFA-GS, an enhanced 3DGS variant that can effectively eliminating floating artifacts. We first conduct theoretical analysis to explore the primary cause of floating artifacts. Based on our analysis, we propose EFA-GS to mitigate floating artifacts. Experiments demonstrate that EFA-GS can effectively eliminate floating artifacts while maintaining fine-grained high-frequency details.

% \bibliographystyle{ACM-Reference-Format}
\bibliographystyle{IEEEbib}
\bibliography{ref}

% \newpage
% %%
% %% If your work has an appendix, this is the place to put it.
% \appendix

\section{Appendix}
\label{sec:Appendix}

\subsection{Specific details about Experiments in Sec.~\ref{sec:expanal}}
\label{sec:A1}

As mentioned in Sec.~\ref{sec:expanal}, we inject noises to Gaussian scales of ship and to Gaussian coordinates of kitchen respectively. In fact, ship does not have SfM initialization priors and it is initialized with random points, so it is meaningless to inject noises to Gaussian coordinates of ship.
In fact, we add noises to the coordinates of ship whereas using original coordinates to execute k-NN algorithm to get initial values of Gaussian scales. This inconsistency is equivalent to injecting noise to Gaussian scales of ship.
We conduct these experiments to illustrate that Gaussians scales and coordinates (corresponding to depth in different camera spaces), two factors related to frequencies and sampling rates according to the Nyquist Sampling algorithm, are important in the optimization process. 

We conduct another quantitative analysis to explore the connection between floaters and under-optimized Gaussians. To explore the effects of under-optimized Gaussians, we control their numbers using the sparse-view setting. Sparse-view causes more under-optimized Gaussians because they are related to the sampling interval, which is view-dependent (Eq. (1) and (2)). Fewer views cause more under-optimized Gaussians. We perform experiments using high-quality initialization on Bonsai (Mip-NeRF360). Results in Tab.~\ref{tab:sparse} indicate that under-optimized Gaussians cause damage (artifacts) to generalizability on test views. Under-optimized Gaussians are more sensitive to noise and easily become artifacts.

We also conduct quantitative analysis to explore the impact of noisy initialization(\(\mathrm{NI}\)). To be specific, we inject Gaussian noise \(n\) with different intensities (\(\mathrm{NI}=\mathrm{Init}+k\cdot n\)) to study its effect. As shown in Tab.~\ref{tab:noisyinit}, stronger noise causes lower PSNR for the baseline. EFA-GS achieves better results, showing robustness to noisy initialization.

\begin{table}[htbp]
    \caption{Sparse-view experiments.}
    \centering
        \begin{tabular}{c|c}
        \toprule
        Mip-splatting & PSNR \\
        \midrule
        8-Views       & 17.38 \\
        12-Views      & 18.44 \\
        16-Views      & 20.74 \\
        \bottomrule
        \end{tabular}
    \label{tab:sparse}
\end{table}

\begin{table}[htbp]
    \caption{Reconstruction results of Bonsai using noisy initializations of different noise intensities.}
    \centering
        \begin{tabular}{c|ccc}
        \toprule
        PSNR & k=1 & k=2 & k=3 \\
        \midrule
        Mip-splatting        & 30.76 & 28.95 & 28.27 \\
        EFA-GS(ours)        & 31.38 & 30.04 & 29.50 \\
        \bottomrule
        \end{tabular}
    \label{tab:noisyinit}
\end{table}

\subsection{Extensive Experiments on Depth-based Regularization (2DGS)}
\label{sec:A2}

\begin{figure}[htbp]
  \caption{Average scalings of Gaussians in training 2DGS on ship dataset~\cite{mildenhall2020nerf}.}
  \centering
  \includegraphics[width=\linewidth]{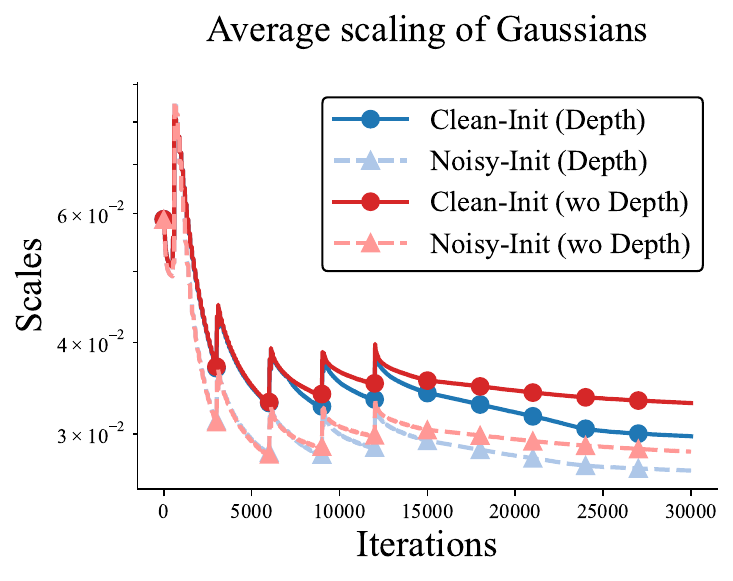}
  \label{fig:aver2dgs}
\end{figure}

To further investigate the limitations of depth-based regularization methods, we perform experiments on 2DGS~\cite{Huang2DGS2024} its depth-dependent loss (depth-distortion loss and normal-consistency loss). To be specific, we first train 2DGS on ship dataset and record the average scaling of Gaussians. We train 2DGS with clean and noisy initialization respectively; then we cancel the depth-related regularization losses of 2DGS and do the same experiments again. Fig.~\ref{fig:aver2dgs} shows that depth-related losses accelerate the process of shrinking Gaussians.  We do not observe artifacts in ship dataset, so we repeat the experiments on kitchen dataset. Experimental results are provided in Tab.~\ref{tab:noisyexp2dgs} and shows that depth-based regularization losses sometimes damage the performance and make 2DGS more sensitive to the noise.

\begin{table}
    \caption{Reconstruction results of kitchen using different initializations. Noisy initialization causes more damages on the rendering quality for testing views than training views, and expands the gap between training and testing views.}
    \centering
    % \resizebox{0.48\textwidth}{!}{
        \begin{tabular}{c|ccc}
        \toprule
        \multirow{2}{*}{PSNR} & \multicolumn{3}{c}{kitchen~\cite{barron2022mip}} \\
                                  & PSNR\(\uparrow\) & SSIM\(\uparrow\) & LPIPS\(\downarrow\) \\
        \midrule
        Clean init (2DGS, with depth)        & 30.36 &  0.92 &  0.13 \\
        Noisy init (2DGS, with depth)        & 28.35 &  0.89 &  0.17 \\
        /Clean - Noisy/ (2DGS, with depth)   &  2.01 &  0.03 &  0.04 \\
        \midrule
        Clean init (2DGS, w/o depth)          & 31.17 &  0.93 &  0.13 \\
        Noisy init (2DGS, w/o depth)          & 29.82 &  0.91 &  0.15 \\
        /Clean - Noisy/ (2DGS, w/o depth)     &  1.35 &  0.02 &  0.02 \\
        \bottomrule
        \end{tabular}
    % }
    \label{tab:noisyexp2dgs}
\end{table}

\subsection{Technical Details about Strategies}
\label{sec:A3}

According to our previous analysis, different regions in a complicated 3D scene have varying sampling rates because of their depths in the pictures, corresponding camera poses and etc. 
It is difficult for low-sampling-rate Gaussians to fit high frequencies well especially when their variances are over-expanded according to Nyquist-Shannon Sampling algorithm. 
Hence, we develop 3 other strategies to mitigating the issue.
\begin{itemize}
\item [1.] To preserve delicate details, EFA-GS performs the LFCF algorithm every few rounds of densification. An hyperparameter \(r\) is set to control the frequency of performing LFCF, and the default value is 2.
\item [2.] To design a coarse-to-fine algorithm, we design an annealing strategy letting all enlarging factors gradually decay to 1 in the training process. To be specific, we let enlarging factors decay exponentially from initial values \(c(i)\) to certain previously set values \(c_{end}\) (default 1).
Given initial value \(c(i)\), The enlarging factor under current iteration is defined as:
\begin{equation}
c = \exp\left(k\cdot x^n\right),
\end{equation}
where \(k=\ln(c(i))\), \(x=\frac{current\_iter-min\_densify\_iter}{max\_densify\_iter-min\_densify\_iter}\). \(n\) is a hyperparameter controlling the decay rate of enlarging factors.

\item [3.] We design a probabilistic strategy to reduce computation costs. It is to assign each Gaussian a probability \(\eta(\cdot)\) to split more Gaussians. Those Gaussians having lower sampling rates would be assigned a higher probability because it is more necessary to reconstruct lower-sampling-rate regions using more Gaussians.
For simplicity, the probability is defined by following formula:
\begin{equation}
\eta(i) = 1 - \theta(i).
\end{equation}

\end{itemize}

Here we provide the ablation studies of these strategies. As shown in Tab.~\ref{tab:ablation2}, regardless of other strategies used, Str 1 and 2 consistently improve performance, while Str 3 reduces computation costs.
\begin{table}[htbp]
    \caption{Ablation studies of 3 strategies on Bonsai.}
    \resizebox{0.48\textwidth}{!}{
    \begin{tabular}{c|ccc|cc}
        \toprule
        EFA-GS Settings          & Str 1 & Str 2 & Str 3 & PSNR\(\uparrow\) & Time\(\downarrow\) \\
        \midrule
        EFA-GS(w/o all str)  &  &  &  & 32.08 & 20'22"\\
        EFA-GS(w/o str 1\&2) &  &  & \(\checkmark\) & 32.12 & 20'12"\\
        EFA-GS(w/o str 1)    &  & \(\checkmark\) & \(\checkmark\) & 32.19 & 20'21"\\
        EFA-GS(w/o str 1\&3) &  & \(\checkmark\) &  & 32.18 & 20'49"\\
        EFA-GS(w/o str 3)    & \(\checkmark\) & \(\checkmark\) &  & 32.50 & 23'10"\\
        EFA-GS(w/o str 2\&3) & \(\checkmark\) &  &  & 32.36 & 22'53"\\
        EFA-GS(w/o str 2)    & \(\checkmark\) &  & \(\checkmark\) & 32.37 & 22'40"\\
        EFA-GS               & \(\checkmark\) & \(\checkmark\) & \(\checkmark\) & 32.56 & 22'55"\\
        \bottomrule
    \end{tabular}
    }
    \label{tab:ablation2}
\end{table}

\subsection{Details about Low-quality Initialization of TanksandTemples Dataset}
\label{sec:A4}
We find a method that produces low-quality SfM initialization without injecting noises. To be specific, we modify several lines of the script convert.py to allow SfM initialization to produce more initial points. This method lower down the performance for the most time.

\subsection{Speed Experiments of EFA-GS on TanksandTemples Dataset}
\label{sec:A5}

We implement EFA-GS on multiple 3DGS variants to evaluate performance and computation costs. As shown in Tab.~\ref{tab:resoftat2}, EFA-GS achieve promising performance improvements on multiple 3DGS variants with limited computation costs. EFA-GS(3DGS) is even faster than Vanilla 3DGS. Here we provide an analysis of this phenomenon:  (1) In LFCF algorithm, gradient checking reuses existing gradients without extra cost. (2) LFCF does not split/clone all Gaussians with gradients larger than the threshold (like the original densification algorithm), which reduces computation. To further evaluate the training speed of EFA-GS, we perform experiments on Bonsai with varying LFCF intervals i (larger i = fewer LFCF operations). As shown in Tab.~\ref{tab:speed}, EFA-GS achieves comparable PSNR with reduced training time.
\begin{table}
    \caption{Reconstruction results on TanksandTemples dataset. }
    \resizebox{0.48\textwidth}{!}{
    \begin{tabular}{c|cccc|cccc}
        \toprule
                      & \multicolumn{4}{c}{Overall} & \multicolumn{4}{c}{Low-Quality Init} \\
                      & PSNR\(\uparrow\) & SSIM\(\uparrow\) & LPIPS\(\downarrow\) & Aver Time(min)\(\downarrow\) & PSNR\(\uparrow\) & SSIM\(\uparrow\) & LPIPS\(\downarrow\) & Aver Time(min)\(\downarrow\)\\
        \midrule
        Vanilla 3DGS  & \cellcolor{Apricot!90}21.51 &  \cellcolor{Apricot!90}0.79 &  \cellcolor{Apricot!90}0.28 & \cellcolor{Apricot!90}32.3  & \cellcolor{Apricot!90}19.11 &  \cellcolor{Apricot!90}0.69 &  \cellcolor{Apricot!90}0.37 & \cellcolor{Apricot!90}31.5 \\
        EFA-GS(3DGS)  & \cellcolor{CoralRed!50}21.69 &  \cellcolor{CoralRed!50}0.80 &  \cellcolor{CoralRed!50}0.28 & \cellcolor{CoralRed!50}29.2  & \cellcolor{CoralRed!50}19.45 &  \cellcolor{CoralRed!50}0.69 &  \cellcolor{CoralRed!50}0.36 & \cellcolor{CoralRed!50}29.0 \\
        \midrule
        eRank-GS      & 18.43 &  0.72 &  0.37 & 75.8  & 16.37 &  0.62 &  0.45 & 72.0 \\
        EFA-GS(eRank) & 20.32 &  0.77 &  0.31 & 80.3  & 17.78 &  0.65 &  0.40 & 77.8 \\
        \midrule
        GOF           & 19.80 &  0.77 &  0.30 & 83.6  & 17.84 &  0.67 &  0.39 & 78.8 \\
        EFA-GS(GOF)   & 20.57 &  0.79 &  0.28 & 86.9  & 19.07 &  0.69 &  0.37 & 85.5 \\
        \bottomrule
    \end{tabular}
    }
    \label{tab:resoftat2}
\end{table}

\begin{table}[htbp]
    \centering
    \caption{Speed Experiment.}
    \begin{tabular}{c|cc}
        \toprule
         & Time\(\downarrow\) & PSNR\(\uparrow\) \\
        \midrule
        Mip-splatting(i=\(\infty\)) & 25'05" & 32.34 \\
        EFA-GS(i=500) & 24'59" & 32.59 \\
        EFA-GS(i=200,default) & 22'55" & 32.56 \\
        EFA-GS(i=100) & 20'21" & 32.19 \\
        \bottomrule
    \end{tabular}
    \label{tab:speed}
\end{table}

\begin{table}[htbp]
    \centering
    \caption{Hyperparameter sensitivity analysis on Mip-NeRF360 dataset. We choose \(c_{max}\) and \(r\) to do the sensitivity analysis.}
    \begin{tabular}{c|ccc}
        \toprule
        PSNR\(\uparrow\) & \(r=1\) & \(r=2\) & \(r=5\) \\
        \midrule
        \(c_{max}=1.50\) & 27.62 & \textbf{27.94} & 27.90  \\
        \(c_{max}=1.75\) & 27.60 & 27.87 & 27.92  \\
        \(c_{max}=2.00\) & 27.58 & 27.89 & 27.91  \\
        \bottomrule
    \end{tabular}
    \label{tab:sense}
\end{table}

\subsection{Qualitative Comaparisons on All Datasets}
\label{sec:A6}
\begin{figure*}[htbp]
  \centering
  \begin{subfigure}{0.135\linewidth}
    \includegraphics[width=\linewidth]{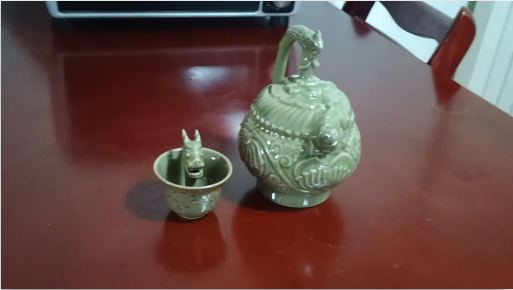}
    \includegraphics[width=\linewidth]{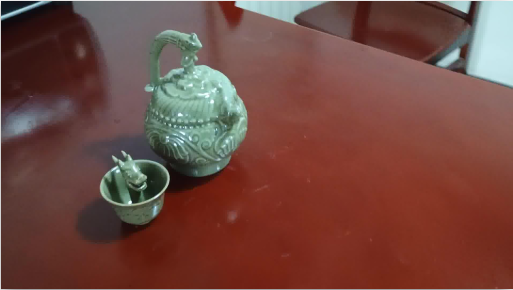}
    \caption{Ground Truth.}
  \end{subfigure}
  \begin{subfigure}{0.135\linewidth}
    \includegraphics[width=\linewidth]{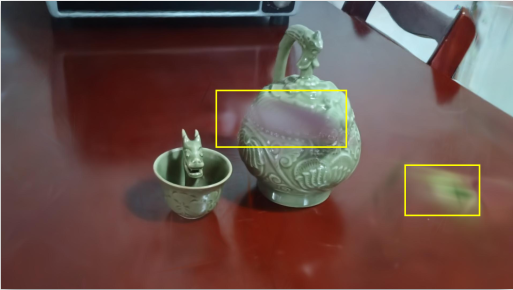}
    \includegraphics[width=\linewidth]{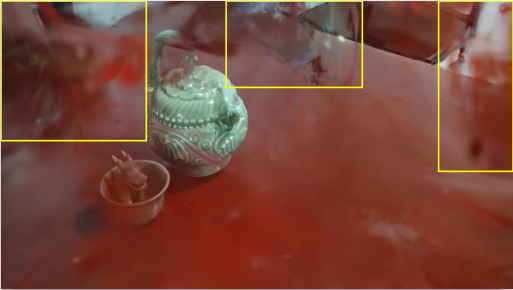}
    \caption{3DGS.}
  \end{subfigure}
  \begin{subfigure}{0.135\linewidth}
    \includegraphics[width=\linewidth]{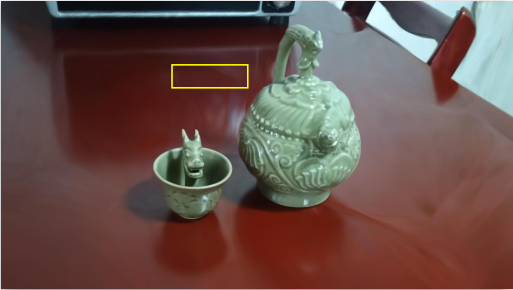}
    \includegraphics[width=\linewidth]{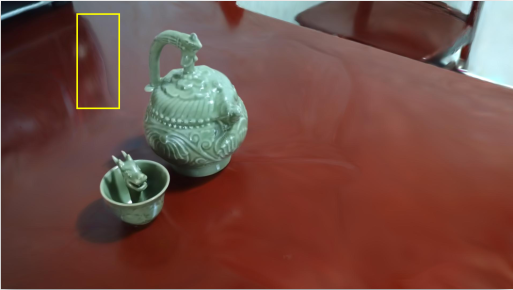}
    \caption{2DGS.}
  \end{subfigure}
  \begin{subfigure}{0.135\linewidth}
    \includegraphics[width=\linewidth]{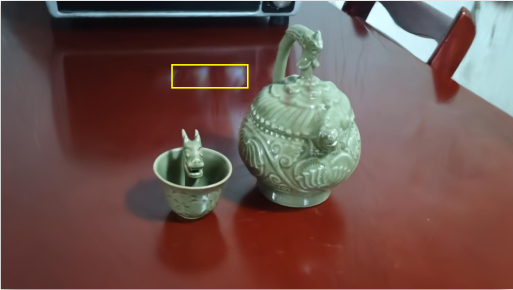}
    \includegraphics[width=\linewidth]{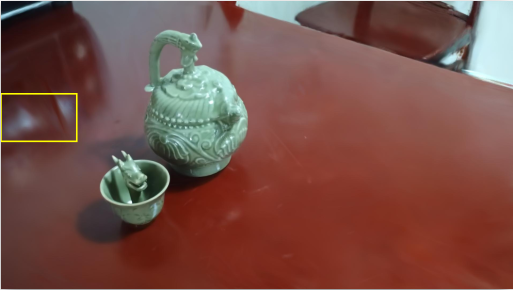}
    \caption{GOF.}
  \end{subfigure}
  \begin{subfigure}{0.135\linewidth}
    \includegraphics[width=\linewidth]{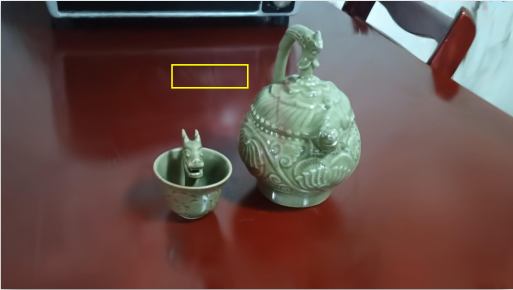}
    \includegraphics[width=\linewidth]{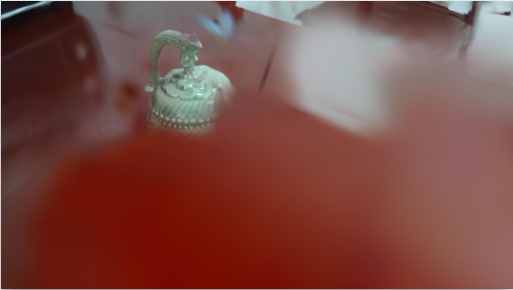}
    \caption{eRank-GS.}
  \end{subfigure}
  \begin{subfigure}{0.135\linewidth}
    \includegraphics[width=\linewidth]{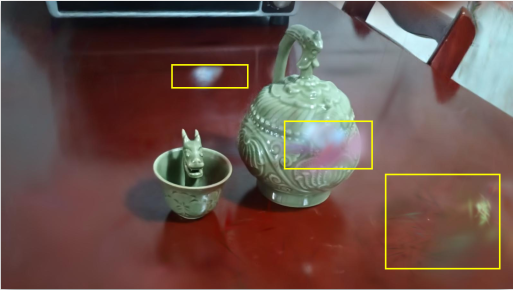}
    \includegraphics[width=\linewidth]{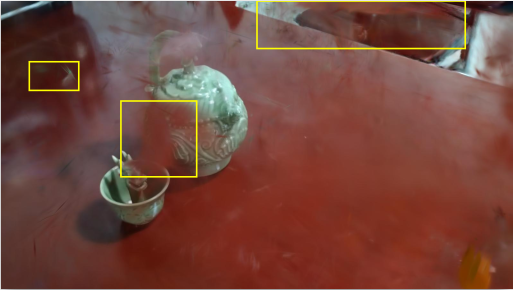}
    \caption{Mip-splatting.}
  \end{subfigure}
  \begin{subfigure}{0.135\linewidth}
    \includegraphics[width=\linewidth]{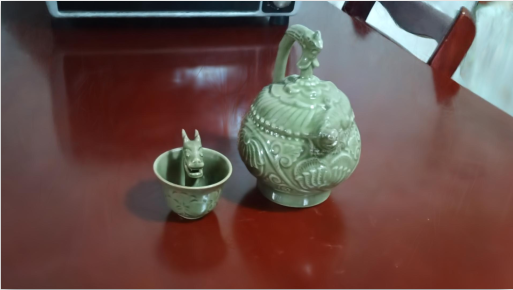}
    \includegraphics[width=\linewidth]{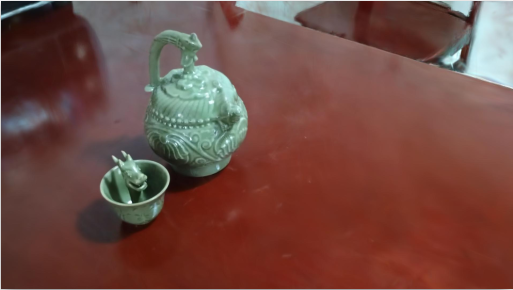}
    \caption{EFA-GS.}
  \end{subfigure}
  \caption{Reconstruction results from different viewpoints and different methods on RWLQ dataset (porcelain). Our EFA-GS achieves state-of-the-art performance among these 3DGS-based methods. EFA-GS effectively eliminates floating artifacts and preserve delicate details while other methods produce floating artifacts or cause erosion of details. This demonstrate the effectiveness of EFA-GS.}
  \label{fig:resofrwlq}
\end{figure*}

\begin{figure*}[htbp]
  \centering
  \begin{subfigure}{0.135\linewidth}
    \includegraphics[width=\linewidth]{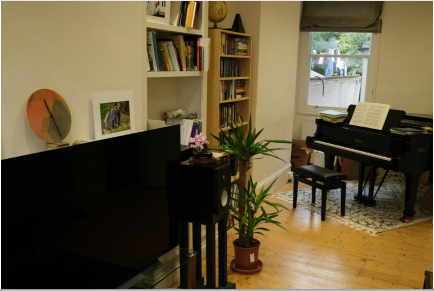}
    \includegraphics[width=\linewidth]{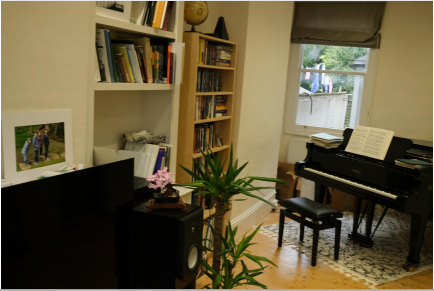}
    \caption{Ground Truth.}
  \end{subfigure}
  \begin{subfigure}{0.135\linewidth}
    \includegraphics[width=\linewidth]{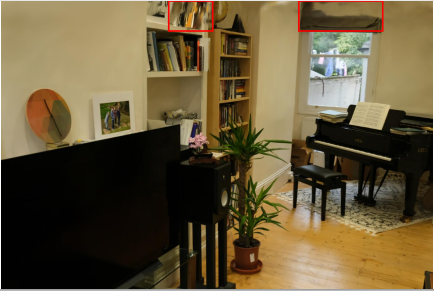}
    \includegraphics[width=\linewidth]{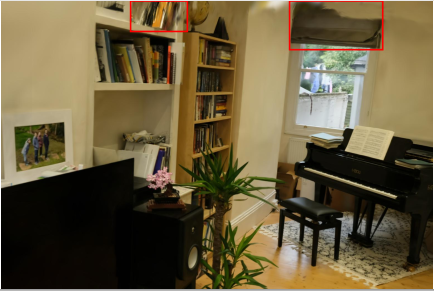}
    \caption{3DGS.}
  \end{subfigure}
  \begin{subfigure}{0.135\linewidth}
    \includegraphics[width=\linewidth]{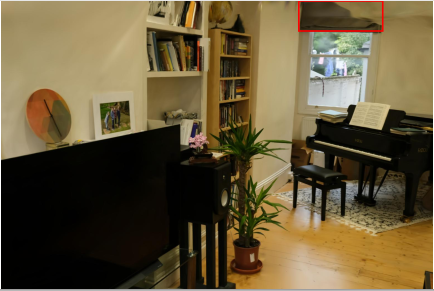}
    \includegraphics[width=\linewidth]{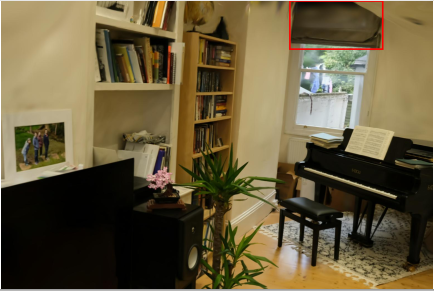}
    \caption{2DGS.}
  \end{subfigure}
  \begin{subfigure}{0.135\linewidth}
    \includegraphics[width=\linewidth]{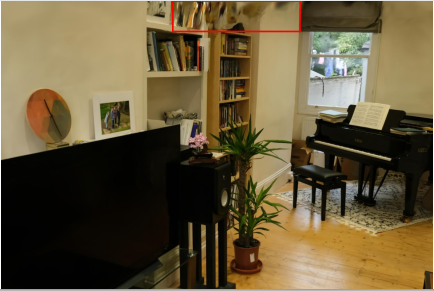}
    \includegraphics[width=\linewidth]{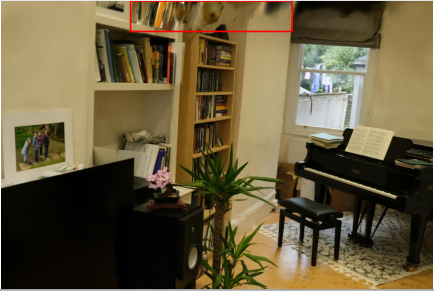}
    \caption{GOF.}
  \end{subfigure}
  \begin{subfigure}{0.135\linewidth}
    \includegraphics[width=\linewidth]{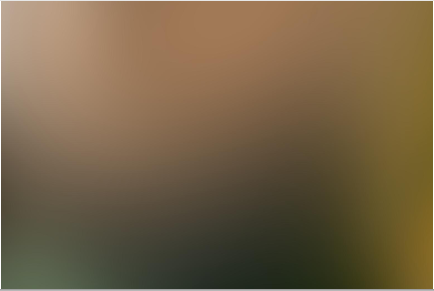}
    \includegraphics[width=\linewidth]{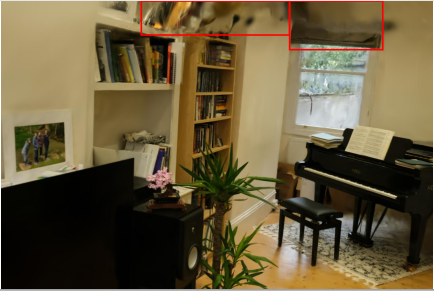}
    \caption{eRank-GS.}
  \end{subfigure}
  \begin{subfigure}{0.135\linewidth}
    \includegraphics[width=\linewidth]{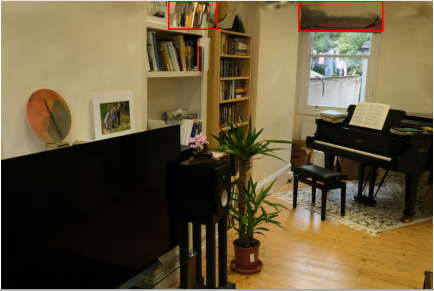}
    \includegraphics[width=\linewidth]{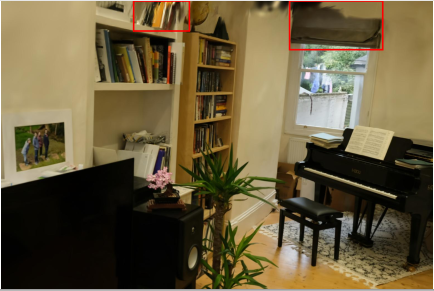}
    \caption{Mip-splatting.}
  \end{subfigure}
  \begin{subfigure}{0.135\linewidth}
    \includegraphics[width=\linewidth]{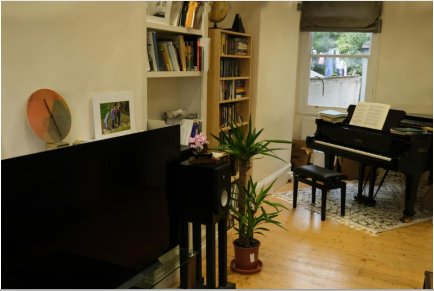}
    \includegraphics[width=\linewidth]{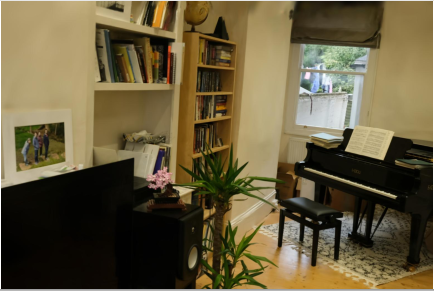}
    \caption{EFA-GS.}
  \end{subfigure}
  \caption{Reconstruction results from different viewpoints and different methods on Mip-NeRF360 dataset (room). Mip-NeRF360 is a high-quality dataset and there are not many floating artifacts. However, EFA-GS still refines some delicate details.}
  \label{fig:resofmip}
\end{figure*}

\begin{figure*}[htbp]
  \centering
  \begin{subfigure}{0.135\linewidth}
    \includegraphics[width=\linewidth]{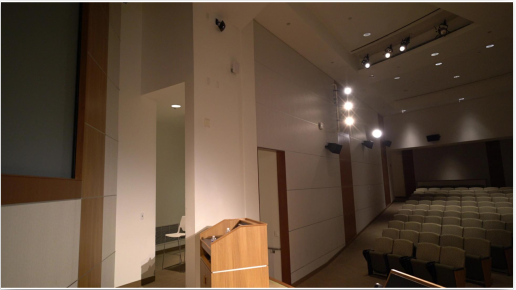}
    \includegraphics[width=\linewidth]{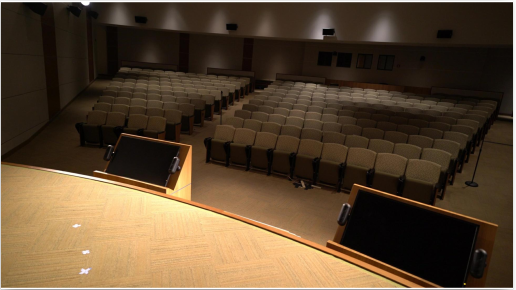}
    \caption{Ground Truth.}
  \end{subfigure}
  \begin{subfigure}{0.135\linewidth}
    \includegraphics[width=\linewidth]{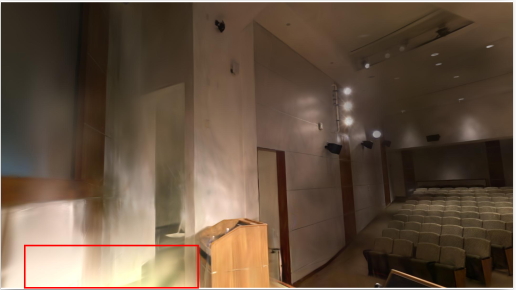}
    \includegraphics[width=\linewidth]{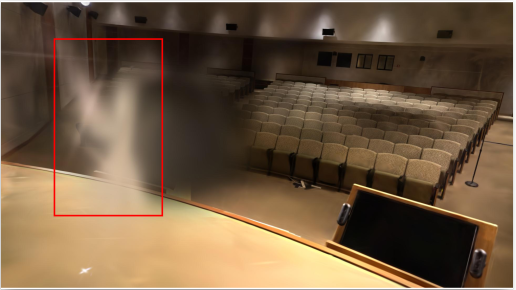}
    \caption{3DGS.}
  \end{subfigure}
  \begin{subfigure}{0.135\linewidth}
    \includegraphics[width=\linewidth]{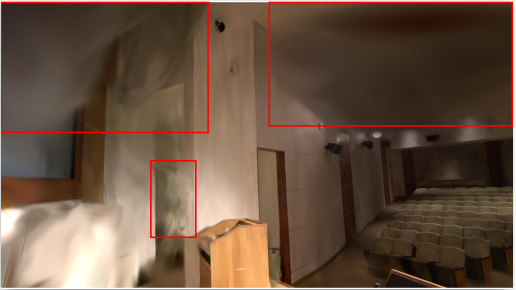}
    \includegraphics[width=\linewidth]{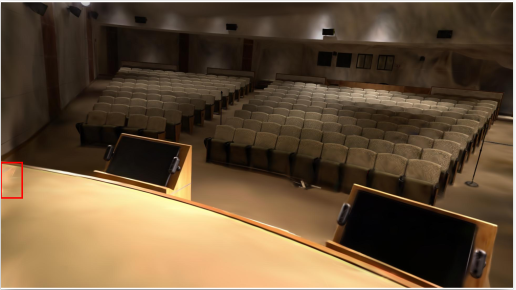}
    \caption{2DGS.}
  \end{subfigure}
  \begin{subfigure}{0.135\linewidth}
    \includegraphics[width=\linewidth]{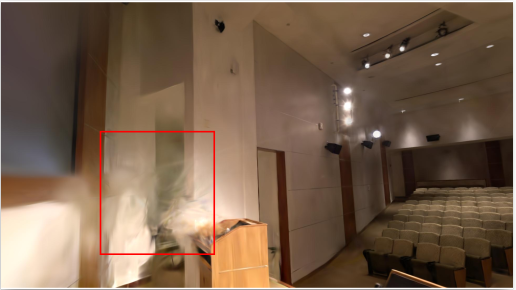}
    \includegraphics[width=\linewidth]{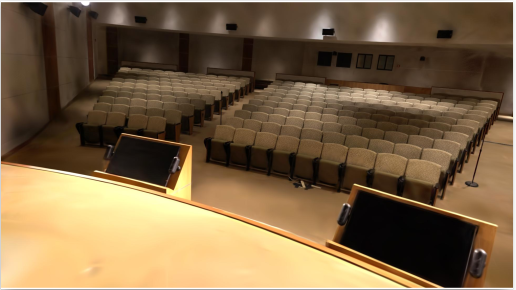}
    \caption{GOF.}
  \end{subfigure}
  \begin{subfigure}{0.135\linewidth}
    \includegraphics[width=\linewidth]{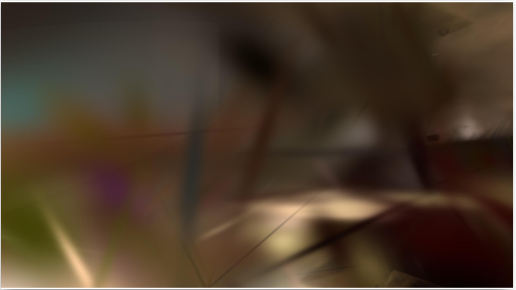}
    \includegraphics[width=\linewidth]{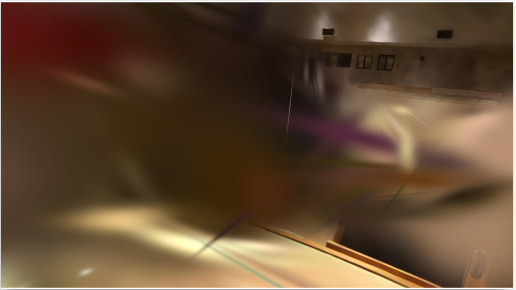}
    \caption{eRank-GS.}
  \end{subfigure}
  \begin{subfigure}{0.135\linewidth}
    \includegraphics[width=\linewidth]{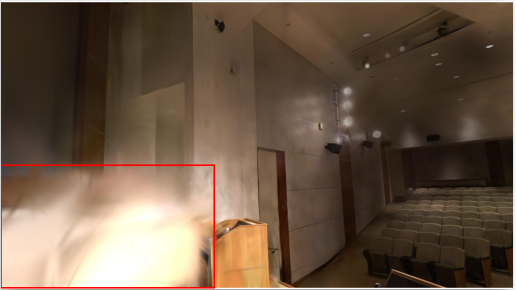}
    \includegraphics[width=\linewidth]{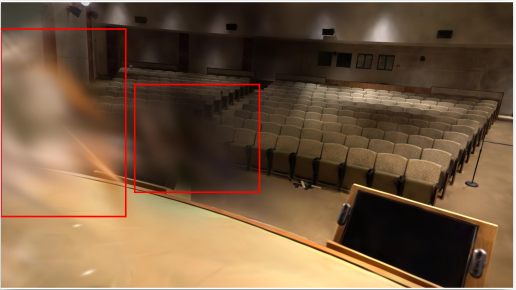}
    \caption{Mip-splatting.}
  \end{subfigure}
  \begin{subfigure}{0.135\linewidth}
    \includegraphics[width=\linewidth]{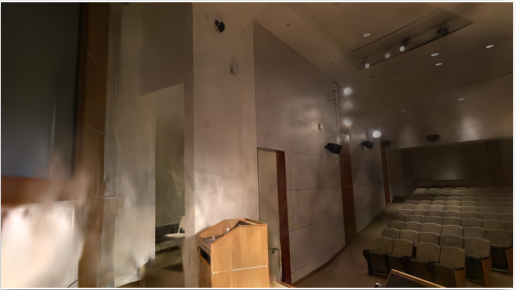}
    \includegraphics[width=\linewidth]{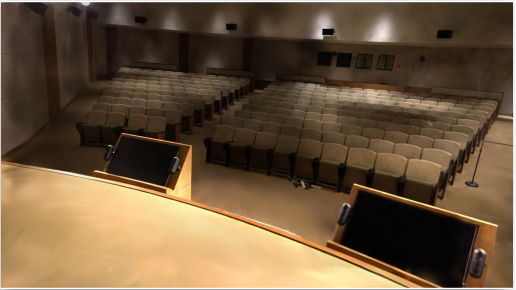}
    \caption{EFA-GS.}
  \end{subfigure}
  \caption{Reconstruction results from different viewpoints and different methods on TanksandTempls dataset (Auditorium). Our EFA-GS also achieves state-of-the-art performance among these 3DGS-based methods. EFA-GS effectively eliminates floating artifacts while other methods produce floating artifacts. This demonstrates the effectiveness of EFA-GS.}
  \label{fig:resoftat}
\end{figure*}

\paragraph*{Qualitative comparison on RWLQ dataset} As shown in Fig.~\ref{fig:resofrwlq}, EFA-GS effectively eliminates artifacts and preserve delicate details at the same time, while other methods either produce floating artifacts or erode details. 2DGS/GOF use depth regularization and they also eliminate floating artifacts. However, the details of the reflective areas on the table are not well preserved.
\paragraph*{Qualitative comparison on Mip-NeRF360 Dataset} As illustrated in Fig.~\ref{fig:resofmip}, even if Mip-NeRF360 is a high-quality dataset and there are not many floating artifacts, EFA-GS still can fix some details in the scene, whereas other methods failed.
\paragraph*{Qualitative comparison on TanksandTemples Dataset}
Although 3DGS achieve the best performance, floating artifacts can still be observed from certain viewpoints, as illustrated in Fig.~\ref{fig:resoftat}. Our EFA-GS mitigates floating artifacts and perserving details in 2 scenes while other methods fail to prevent floating artifacts from seriously damaging render quality.

\subsection{Hyperparameter Sensitivity Analysis}
\label{sec:A7}
To sufficiently evaluate the robustness of EFA-GS, we also conduct experiments of EFA-GS under different hyperparameter settings. We choose 2 major hyperparameters \(c_{max}\) and \(r\) to do the sensitivity analysis. The default setting is \(c_{max}=1.5, r=2\). The dataset we use is Mip-NeRF360 and the evaluation metric is PSNR. We conduct experiments on all 9 scenes of Mip-NeRF360 dataset and report the average performance. 

As shown in Tab.~\ref{tab:sense}, the default setting achieve the best performance among others.

\subsection{Full Reconstruction Results}
\label{sec:A8}
In this section, we provide full experimental results on the 3 datasets.

Full experiments on RWLQ dataset are provided in Tab.~\ref{tab:resofrwlqpsnr}, \ref{tab:resofrwlqssim} and \ref{tab:resofrwlqlpips}.
\begin{table}
    \caption{Reconstruction results on RWLQ dataset (PSNR). }
    \begin{tabular}{c|cccc}
        \toprule
        PSNR\(\uparrow\) & astronaut & porcelain & specimen & tank \\
        \midrule
        Vanilla 3DGS  & 28.19 & 26.65 & 26.58 & 29.41 \\
        2DGS          & 27.43 & 28.52 & 26.66 & 29.37 \\
        GOF           & 27.43 & 28.15 & 26.48 & 29.62 \\
        eRank-GS      & 19.35 & 23.04 & 23.25 & 27.00 \\
        Mip-splatting & 26.68 & 25.89 & 25.06 & 29.02 \\
        \midrule
        EFA-GS(ours)  & 28.79 & 29.46 & 26.29 & 28.84 \\
        \bottomrule
    \end{tabular}
    \label{tab:resofrwlqpsnr}
\end{table}

\begin{table}
    \caption{Reconstruction results on RWLQ dataset (SSIM). }
    \begin{tabular}{c|cccc}
        \toprule
        SSIM\(\uparrow\) & astronaut & porcelain & specimen & tank \\
        \midrule
        Vanilla 3DGS  &  0.98 &  0.93 &  0.92 &  0.97 \\
        2DGS          &  0.97 &  0.94 &  0.92 &  0.97 \\
        GOF           &  0.97 &  0.94 &  0.92 &  0.97 \\
        eRank-GS      &  0.93 &  0.90 &  0.87 &  0.96 \\
        Mip-splatting &  0.97 &  0.92 &  0.91 &  0.96 \\
        \midrule
        EFA-GS(ours)  &  0.97 &  0.94 &  0.91 &  0.96 \\
        \bottomrule
    \end{tabular}
    \label{tab:resofrwlqssim}
\end{table}

\begin{table}
    \caption{Reconstruction results on RWLQ dataset (LPIPS). }
    \begin{tabular}{c|cccc}
        \toprule
        LPIPS\(\downarrow\) & astronaut & porcelain & specimen & tank \\
        \midrule
        Vanilla 3DGS  &  0.11 &  0.21 &  0.19 &  0.10 \\
        2DGS          &  0.11 &  0.20 &  0.19 &  0.10 \\
        GOF           &  0.12 &  0.20 &  0.19 &  0.11 \\
        eRank-GS      &  0.14 &  0.24 &  0.22 &  0.11 \\
        Mip-splatting &  0.11 &  0.22 &  0.20 &  0.11 \\
        \midrule
        EFA-GS(ours)  &  0.11 &  0.20 &  0.19 &  0.11 \\
        \bottomrule
    \end{tabular}
    \label{tab:resofrwlqlpips}
\end{table}

Full experiments on Mip-NeRF360 dataset are provided in Tab.~\ref{tab:resofmippsnr}, \ref{tab:resofmipssim} and \ref{tab:resofmiplpips}.

\begin{table}
    \caption{Reconstruction results on Mip-NeRF360 dataset (PSNR). }
    \resizebox{0.48\textwidth}{!}{
    \begin{tabular}{c|ccccccccc}
        \toprule
        PSNR\(\uparrow\) & bicycle & bonsai & counter & flowers & garden & kitchen & room & stump & treehill \\
        \midrule
        Vanilla 3DGS  & 25.39 & 32.26 & 28.96 & 21.89 & 27.28 & 31.43 & 31.75 & 26.60 & 22.69\\
        2DGS          & 25.15 & 31.30 & 28.18 & 21.36 & 27.02 & 30.36 & 30.62 & 26.38 & 22.58\\
        GOF           & 25.46 & 31.58 & 28.69 & 21.69 & 27.42 & 30.75 & 30.98 & 26.99 & 22.44\\
        Mip-splatting & 25.96 & 32.34 & 29.29 & 22.01 & 28.04 & 31.91 & 31.94 & 27.16 & 22.60\\
        \midrule
        EFA-GS(ours)  & 25.86 & 32.56 & 29.36 & 21.98 & 28.02 & 31.81 & 32.13 & 27.11 & 22.62\\
        \bottomrule
    \end{tabular}
    }
    \label{tab:resofmippsnr}
\end{table}

\begin{table}
    \caption{Reconstruction results on Mip-NeRF360 dataset (SSIM). }
    \resizebox{0.48\textwidth}{!}{
    \begin{tabular}{c|ccccccccc}
        \toprule
        SSIM\(\uparrow\) & bicycle & bonsai & counter & flowers & garden & kitchen & room & stump & treehill \\
        \midrule
        Vanilla 3DGS  &  0.76 &  0.95 &  0.91 &  0.62 &  0.87 &  0.93 &  0.93 &  0.77 &  0.65\\
        2DGS          &  0.75 &  0.94 &  0.90 &  0.59 &  0.85 &  0.92 &  0.91 &  0.77 &  0.64\\
        GOF           &  0.79 &  0.94 &  0.90 &  0.64 &  0.87 &  0.92 &  0.92 &  0.79 &  0.64\\
        Mip-splatting &  0.80 &  0.95 &  0.92 &  0.66 &  0.88 &  0.94 &  0.93 &  0.80 &  0.66\\
        \midrule
        EFA-GS(ours)  &  0.80 &  0.95 &  0.92 &  0.65 &  0.88 &  0.93 &  0.93 &  0.80 &  0.66\\
        \bottomrule
    \end{tabular}
    }
    \label{tab:resofmipssim}
\end{table}

\begin{table}
    \caption{Reconstruction results on Mip-NeRF360 dataset (LPIPS). }
    \resizebox{0.48\textwidth}{!}{
    \begin{tabular}{c|ccccccccc}
        \toprule
        LPIPS\(\downarrow\) & bicycle & bonsai & counter & flowers & garden & kitchen & room & stump & treehill \\
        \midrule
        Vanilla 3DGS  &  0.22 &  0.17 &  0.18 &  0.33 &  0.11 &  0.12 &  0.19 &  0.22 &  0.32\\
        2DGS          &  0.26 &  0.19 &  0.21 &  0.36 &  0.14 &  0.13 &  0.22 &  0.25 &  0.37\\
        GOF           &  0.18 &  0.20 &  0.20 &  0.28 &  0.11 &  0.14 &  0.22 &  0.20 &  0.30\\
        Mip-splatting &  0.16 &  0.16 &  0.17 &  0.27 &  0.09 &  0.11 &  0.18 &  0.18 &  0.27\\
        \midrule
        EFA-GS(ours)  &  0.17 &  0.16 &  0.17 &  0.28 &  0.09 &  0.11 &  0.18 &  0.19 &  0.28\\
        \bottomrule
    \end{tabular}
    }
    \label{tab:resofmiplpips}
\end{table}

Full experiments on TanksandTemples dataset are provided in Tab.~\ref{tab:resoftatpsnr}, \ref{tab:resoftatssim} and \ref{tab:resoftatlpips}.

\begin{table}
    \caption{Reconstruction results on TanksandTemples dataset (PSNR). }
    \resizebox{0.48\textwidth}{!}{
    \begin{tabular}{c|cccccc}
        \toprule
        PSNR\(\uparrow\) & Auditorium & Ballroom & Courtroom & Museum & Palace & Temple \\
        \midrule
        Vanilla 3DGS  & 24.17 & 20.04 & 23.26 & 21.31 & 19.68 & 20.63\\
        2DGS          & 23.89 & 19.21 & 23.02 & 21.51 & 18.90 & 20.46\\
        GOF           & 22.44 & 17.74 & 21.44 & 20.19 & 16.54 & 20.44\\
        eRank-GS      & 23.17 & 15.93 & 21.47 & 20.87 & 14.40 & 14.72\\
        Mip-splatting & 23.79 & 18.86 & 22.69 & 20.48 & 17.91 & 20.09\\
        \midrule
        EFA-GS(ours)  & 24.44 & 19.71 & 23.05 & 20.85 & 19.54 & 20.22\\
        \bottomrule
    \end{tabular}
    }
    \label{tab:resoftatpsnr}
\end{table}

\begin{table}
    \caption{Reconstruction results on TanksandTemples dataset (SSIM). }
    \resizebox{0.48\textwidth}{!}{
    \begin{tabular}{c|cccccc}
        \toprule
        SSIM\(\uparrow\) & Auditorium & Ballroom & Courtroom & Museum & Palace & Temple \\
        \midrule
        Vanilla 3DGS  &  0.88 &  0.72 &  0.80 &  0.79 &  0.76 &  0.81\\
        2DGS          &  0.87 &  0.69 &  0.79 &  0.79 &  0.74 &  0.80\\
        GOF           &  0.88 &  0.66 &  0.78 &  0.78 &  0.72 &  0.81\\
        eRank-GS      &  0.85 &  0.57 &  0.76 &  0.77 &  0.67 &  0.69\\
        Mip-splatting &  0.88 &  0.68 &  0.79 &  0.78 &  0.73 &  0.81\\
        \midrule
        EFA-GS(ours)  &  0.89 &  0.72 &  0.80 &  0.79 &  0.76 &  0.81\\
        \bottomrule
    \end{tabular}
    }
    \label{tab:resoftatssim}
\end{table}

\begin{table}
    \caption{Reconstruction results on TanksandTemples dataset (LPIPS). }
    \resizebox{0.48\textwidth}{!}{
    \begin{tabular}{c|cccccc}
        \toprule
        LPIPS\(\downarrow\) & Auditorium & Ballroom & Courtroom & Museum & Palace & Temple \\
        \midrule
        Vanilla 3DGS  &  0.27 &  0.30 &  0.25 &  0.24 &  0.35 &  0.29\\
        2DGS          &  0.31 &  0.35 &  0.30 &  0.26 &  0.38 &  0.31\\
        GOF           &  0.27 &  0.36 &  0.27 &  0.23 &  0.40 &  0.28\\
        eRank-GS      &  0.31 &  0.45 &  0.32 &  0.27 &  0.47 &  0.41\\
        Mip-splatting &  0.27 &  0.34 &  0.25 &  0.24 &  0.37 &  0.29\\
        \midrule
        EFA-GS(ours)  &  0.26 &  0.29 &  0.24 &  0.24 &  0.35 &  0.28\\
        \bottomrule
    \end{tabular}
    }
    \label{tab:resoftatlpips}
\end{table}

\end{document}